\documentclass[11pt]{article}

\usepackage[final]{acl}
\usepackage{times}
\usepackage{latexsym}
\usepackage{subcaption}
\usepackage{url}
\usepackage{booktabs}
\usepackage{graphicx}
\usepackage{multirow}
\usepackage{makecell}
\usepackage{wrapfig}
\usepackage[table]{xcolor}
\usepackage{pifont} 
\newcommand{\cmark}{\ding{51}}%
\newcommand{\xmark}{\ding{55}}%
\usepackage{colortbl}
\usepackage{soul}
\usepackage{url}
\usepackage[utf8]{inputenc}
\usepackage{graphicx}
\usepackage{amsmath}
\usepackage{amsthm}
\usepackage{booktabs}
\usepackage{algorithm}
\usepackage{algorithmic}
\usepackage[switch]{lineno}

\usepackage[T1]{fontenc}
\usepackage[utf8]{inputenc}

\usepackage{microtype}

\usepackage{inconsolata}

\usepackage{graphicx}

\title{LifeAgentBench: Benchmarking LLMs for Long-Horizon, Cross-Dimensional Lifestyle Health Reasoning}

\author{
\begin{tabular}{ccccc}
Ye Tian$^{1}$\thanks{Ye Tian and Zihao Wang contributed equally.} &
Zihao Wang$^{1}$\footnotemark[1] &
Onat Gungor$^{1,3}$ &
Xiaoran Fan$^{2}$ &
Tajana Rosing$^{1}$
\end{tabular}
\\
$^{1}$University of California San Diego, La Jolla, CA, USA \\
$^{2}$Google Research, Mountain View, CA, USA \\
$^{3}$West Virginia University, Morgantown, WV, USA \\
\texttt{\{yet002,ziw140,tajana\}@ucsd.edu}\quad
\texttt{onat.gungor@mail.wvu.edu} \quad
\texttt{vanxf@google.com} \\
\url{https://gdfwj.github.io/LifeAgentBench/}
}

\newcommand{\systemname}{{LifeAgentBench}~}
\newcommand{\systemnamenospace}{{LifeAgentBench}}

\newcommand{\agentmname}{{LifeAgent}~}
\newcommand{\agentmnamenospace}{{LifeAgent}}

\newcommand{\FQ}{{Fact Query}}
\newcommand{\AS}{{Aggregated Statistics}}
\newcommand{\NC}{{Numeric comparison}}
\newcommand{\CQ}{{Conditional Query}}
\newcommand{\TA}{{Trend Analysis}}

\usepackage{xcolor}
\usepackage{tabularx}
\usepackage{hyperref}
\usepackage{placeins}

\begin{document}
\maketitle
\begin{abstract}
Personalized lifestyle health analysis requires long-horizon, multi-dimensional reasoning over heterogeneous lifestyle signals, and recent advances in mobile sensing and large language models (LLMs) make such support increasingly feasible.
However, the capabilities of current LLMs in this setting remain insufficiently understood due to the lack of systematic benchmarks.
In this paper, we introduce LifeAgentBench, a large-scale QA benchmark for long-horizon, cross-dimensional, and multi-user lifestyle health reasoning, containing 22,573 questions spanning from basic retrieval to complex reasoning.
We release an extensible benchmark construction pipeline and a standardized evaluation protocol, deriving verifiable answers through executable queries and programs to support reliable assessment.
We then systematically evaluate 13 representative LLMs on LifeAgentBench and identify key bottlenecks in long-horizon aggregation and cross-dimensional reasoning.
Motivated by these findings, we propose LifeAgent, a tool-augmented reasoning baseline that decomposes complex queries, performs multi-step evidence retrieval, and invokes tools for deterministic aggregation.
LifeAgent substantially enhances LLMs' capabilities on challenging reasoning tasks, achieving clear improvements over widely used baselines and showing potential for health reasoning in everyday scenarios.
The benchmark is publicly available.
\end{abstract}
%key words:Large Language Models, Question Answering, Benchmark, Lifestyle Health Reasoning
%\footnote{\url{https://anonymous.4open.science/r/LifeAgentBench-CE7B}}

\begin{figure}[t]
\begin{center}
    \includegraphics[width=0.86\linewidth]{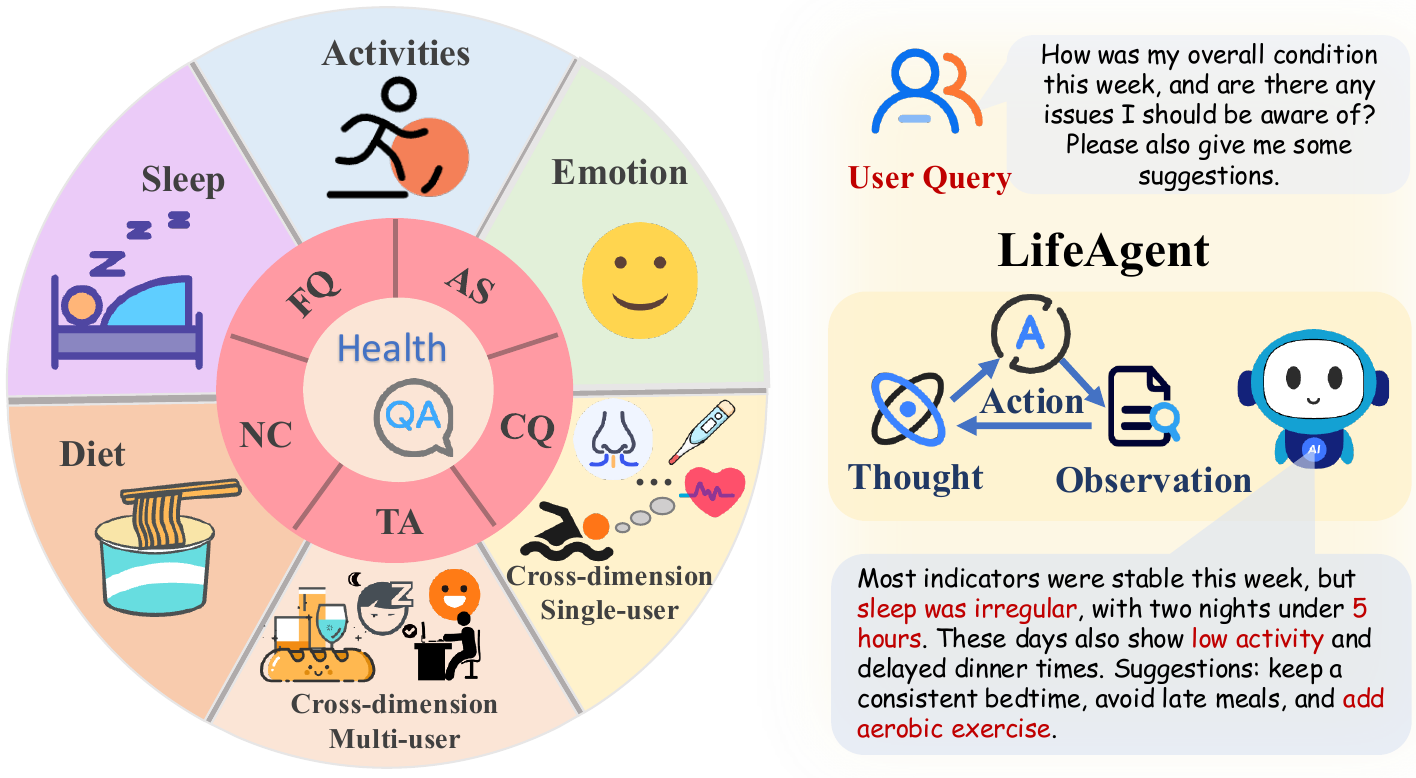}
\end{center}
\vspace{-0.4cm}
\caption{\systemnamenospace, a benchmark for long-horizon, cross-dimensional lifestyle health reasoning, together with \agentmnamenospace, a reasoning baseline.}
\label{fig:Firstpicture}
\vspace{-0.4cm}
\end{figure}

\section{Introduction}
Transforming everyday lifestyle signals into timely and personalized analysis and feedback can support healthier living and help reduce disease burden~\cite{hietbrink2025systematic}.
Such support is closely connected to major lifestyle factors, as physical inactivity, insufficient sleep, unhealthy diets, and psychological stress are well-established risk factors for the onset and progression of noncommunicable diseases~\cite{WHO_NCDs,vaccarino2025mental}.
Recent advances in wearables and mobile health applications have enabled continuous and fine-grained monitoring of daily-life signals, including physical activity, sleep patterns, dietary intake, and stress-related indicators~\cite{jamieson2025guide,Apple_Watch,Google_PixelWatch4,oei2024image,zareeia2025classification,mcduff2025evidence}.
These longitudinal and multi-dimensional lifestyle records provide a solid foundation for personalized health-related analysis.
%but generating reliable feedback from them requires reasoning over heterogeneous signals across long time horizons, identifying relevant evidence, and synthesizing information across multiple lifestyle dimensions.

Reasoning over heterogeneous lifestyle signals across long time spans and producing reliable feedback is challenging.
In daily life, users often prefer natural-language interaction over directly inspecting raw lifelog records.
They may ask questions such as ``How has my sleep quality been over the past week?'' or ``When I increase aerobic exercise, does my deep sleep also increase?''
Answering these questions requires aggregating measurements over extended periods, comparing different periods or conditions, and relating signals across different lifestyle dimensions.
For example, a single night of reduced sleep may be negligible in isolation, but when combined with rising stress and an irregular diet, it may indicate risks that single-dimensional analysis would miss.
Existing machine learning models are effective for narrow tasks such as classification and prediction, but they provide limited support for holistic, user-facing reasoning and explanation over multi-domain lifestyle factors~\cite{abdelaal2024exploring}.

Large language models (LLMs) have shown strong capabilities across diverse domains, offering new opportunities for natural-language analysis and feedback over personal health-related records~\cite{bedi2025testing}.
Recent studies have begun to explore LLMs for specific health and lifestyle applications, such as sleep assessment~\cite{khasentino2025personal}, activity analysis~\cite{yu2025sensorchat}, daily log generation~\cite{tian2025dailyllm}, and emotion analysis~\cite{xu2024mental}.
Several health and lifestyle QA benchmarks have also been introduced, covering nutritional decision making (NGQA~\cite{zhang2024ngqa}), activity analysis (SensorQA~\cite{reichman2025sensorqa}), sleep health (SleepQA~\cite{bojic2022sleepqa}), emotional support (MentalChat16K~\cite{xu2025mentalchat16k}), and lifelog analysis (OpenLifelogQA~\cite{tran2025openlifelogqa}).
However, existing benchmarks and evaluations often focus on single domains, short temporal contexts, or task-specific applications.
As a result, it still lacks a unified benchmark and evaluation protocol for systematically assessing LLMs under long-horizon, cross-dimensional, and multi-user lifestyle health reasoning.

To fill this gap, we introduce \systemnamenospace, a large-scale QA benchmark for evaluating LLMs on long-horizon, cross-dimensional lifestyle health reasoning, as shown in Figure~\ref{fig:Firstpicture}.
Built from multi-dimensional lifestyle records, \systemname contains 22,573 questions spanning basic fact retrieval, aggregation-intensive reasoning, cross-dimensional analysis, and multi-user comparison.
To support reliable and scalable evaluation, we develop an extensible benchmark construction pipeline that aligns lifelog records into a relational database, instantiates questions from task templates, and derives deterministic ground-truth answers through executable queries and programs.
This design enables scalable dataset construction while grounding each answer in verifiable evidence.

With \systemnamenospace, we evaluate 13 representative LLMs under two common evaluation settings and analyze their performance and failure modes: \textit{Context Prompting}~\cite{sui2024table} and \textit{Database-augmented Prompting}~\cite{biswal2024text2sql}. The results reveal major bottlenecks in long-horizon aggregation and cross-dimensional, multi-user reasoning, suggesting that tool-augmented grounding and deterministic aggregation are critical for reliable lifestyle health reasoning.
Motivated by these findings, we design \agentmname~as a strong baseline that decomposes complex queries, performs multi-step evidence retrieval, and invokes tools to handle aggregation-intensive reasoning over long time horizons.
Experiments show that \agentmname~substantially improves performance on the most challenging cross-dimensional and multi-user questions, and a realistic case study further demonstrates its potential for supporting everyday lifestyle health reasoning.
In summary, our main contributions are as follows:
\begin{itemize}
    \item We introduce \systemnamenospace, a large-scale benchmark for long-horizon, cross-dimensional, and multi-user lifestyle health reasoning over personal lifestyle records.

    \item We systematically evaluate 13 representative LLMs and reveal their key bottlenecks in evidence retrieval, long-window aggregation, and cross-dimensional reasoning.

    \item We propose \agentmnamenospace, a tool-augmented baseline that enhances LLMs' reasoning capabilities on challenging tasks through multi-step evidence retrieval and tool use.
\end{itemize}

\begin{table*}[h]
\centering
\small
\resizebox{\textwidth}{!}{\begin{tabular}{llllcc}
\toprule
Method & Task & Scale  & Covered Dimensions & Multi-user & Cross-dimension \\
\midrule
NGQA~\cite{zhang2024ngqa} & Nutrition reasoning & 13.8K & Nutrition Health & \xmark & \xmark \\
SensorQA~\cite{reichman2025sensorqa} & Daily-life QA & 5.6K  & Activity and Location & \xmark & \xmark \\
SleepQA~\cite{bojic2022sleepqa} & Sleep guidance & 7K  & Sleep data & \xmark & \xmark \\
MentalChat16K~\cite{xu2025mentalchat16k} & Mental health dialogue & 16.1K  & Emotion / Mental health & \xmark & \xmark \\
OpenLifelogQA~\cite{tran2025openlifelogqa} & Lifelog QA & 14.2K  & Multi-modal lifestyle & \xmark & \cmark \\
\cmidrule(lr){1-6}
\textbf{\systemname (Ours)} & \textbf{Lifestyle  health QA} & \textbf{22.6K}  & Diet, activity, sleep, and emotion data & \cmark & \cmark \\
\bottomrule
\end{tabular}}
\caption{Comparison of \systemname with existing health and lifestyle QA benchmarks.}\label{tab:Related work comparison}
\end{table*}
%\footnotesize{(\textit{Multi-user} and \textit{Cross-dimension} indicate whether the dataset supports multi-user reasoning and cross-dimensional QA reasoning.)}

%%%
\begin{figure*}[t]
\centering
\begin{subfigure}{0.19\linewidth}
    \centering
    \includegraphics[width=\linewidth]{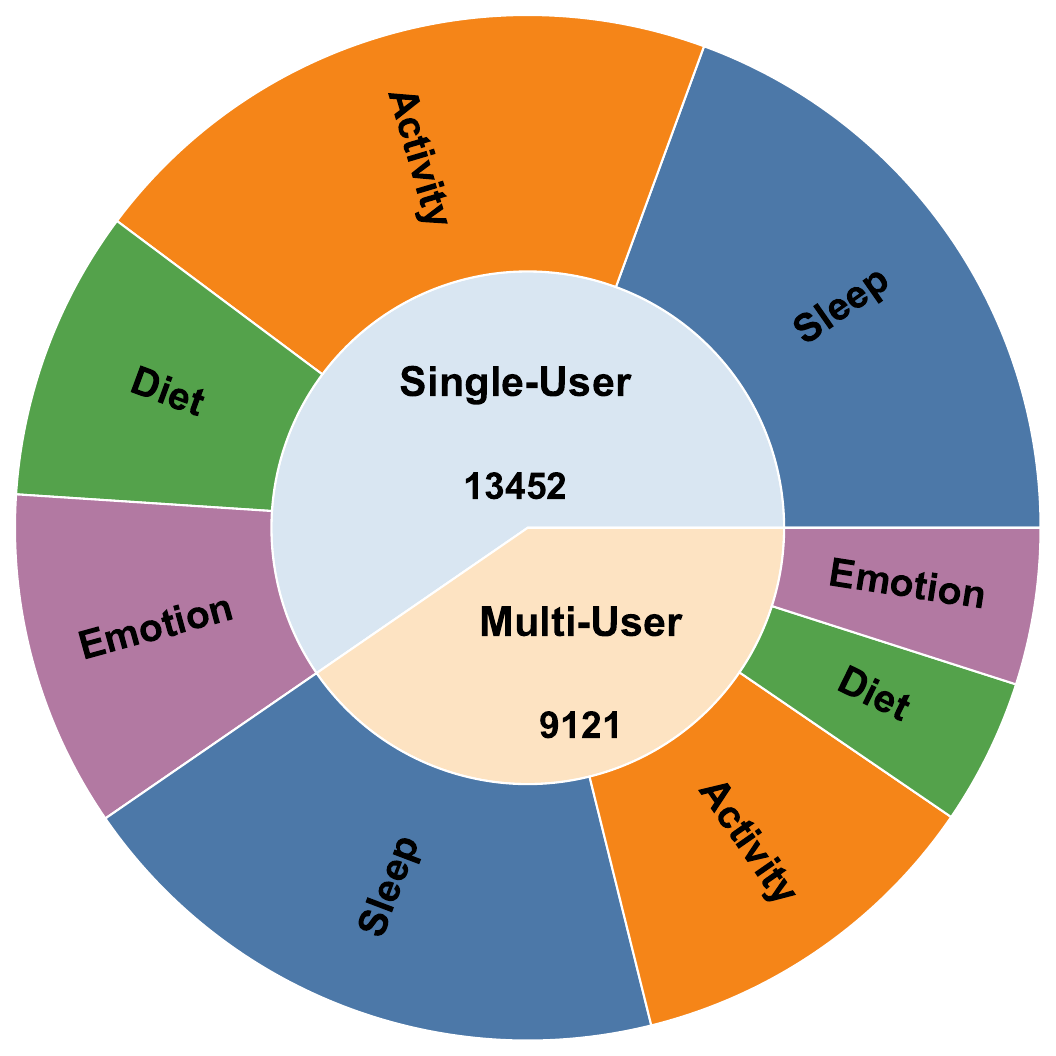}
    \caption{}
    \label{fig:SingleANDmulti}
\end{subfigure}
%\hfill
\quad \quad
\begin{subfigure}{0.32\linewidth}
    \centering
\includegraphics[width=\linewidth]{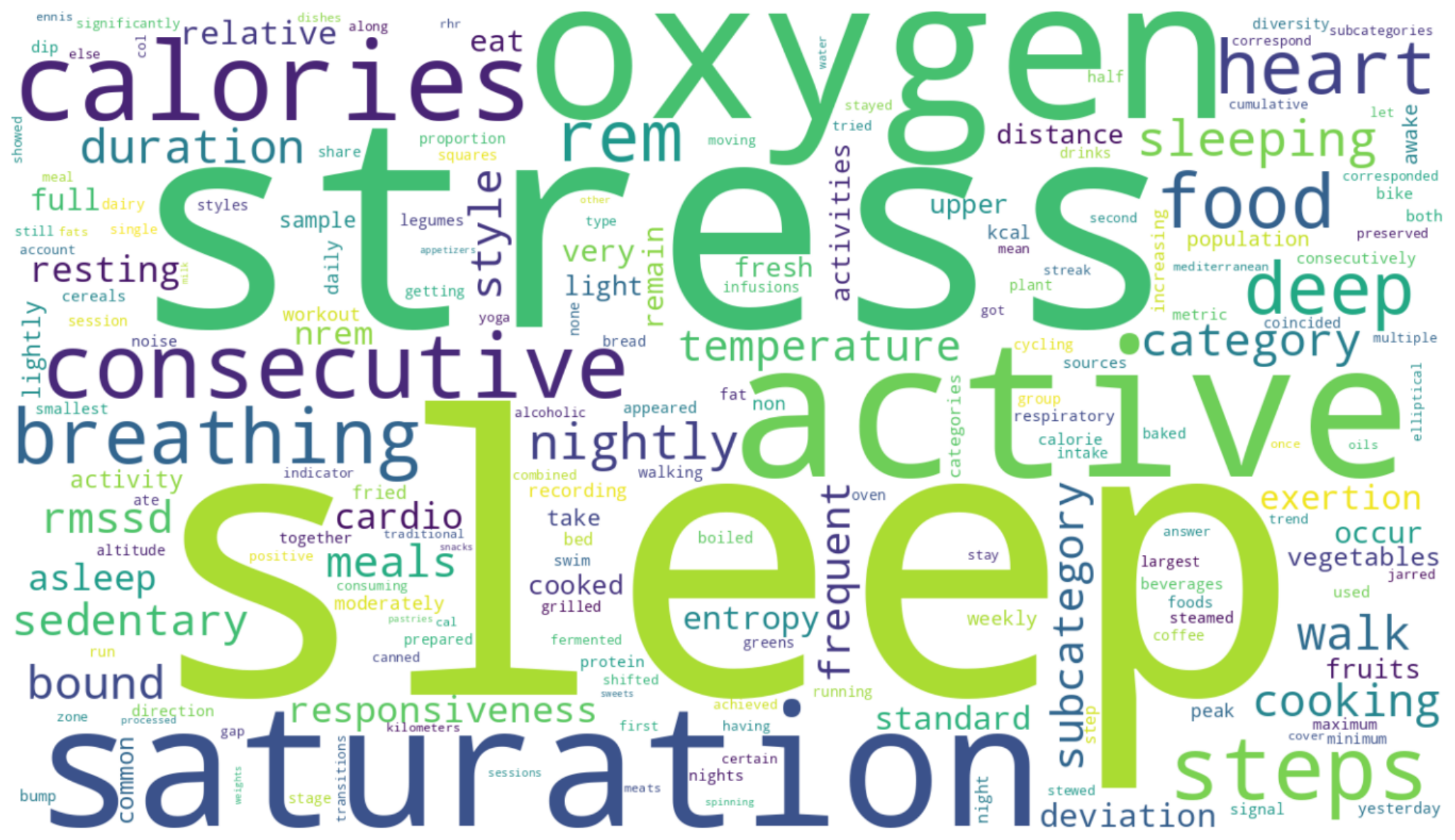}
    \caption{}
\label{fig:Question_wordcloud}
\end{subfigure}
\quad \quad
\begin{subfigure}
{0.19\linewidth}
    \centering
\includegraphics[width=\linewidth]{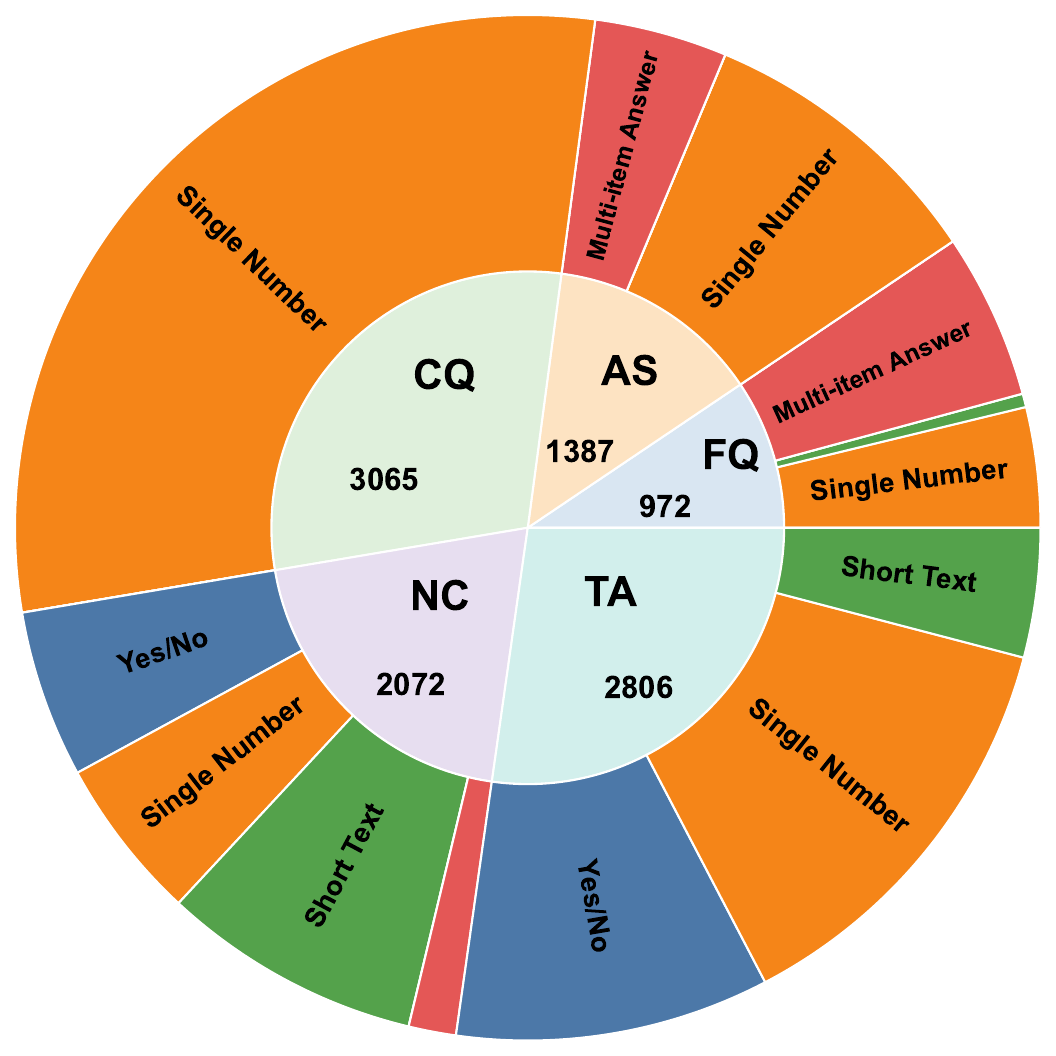}
    \caption{}
    \label{fig:QAcategory}
\end{subfigure}
\vspace{-0.30cm}
\caption{Overview statistics of \systemnamenospace: (a) domain distribution for single-user and multi-user questions, (b) question word-frequency visualization, and (c) distributions of question types and answer formats.}
\label{fig:allthree}
\vspace{-0.4cm}
\end{figure*}

\section{Related Work}
\textbf{Lifestyle Data for Health Analysis.}
Lifestyle datasets provide essential resources for modeling daily health reasoning. 
Early resources are often short-span or single-focus, such as MMASH~\cite{rossi2020multilevel} for sleep and psychological analysis, WESAD~\cite{philip2018multimodal} for stress and affective states, and CAPTURE-24~\cite{chan2024capture} for large-scale activity recognition with sleep diaries. 
More recent lifelogging efforts extend to longer-term and more diverse sensing, including LifeSnaps~\cite{yfantidou2022lifesnaps}, GLOBEM~\cite{xu2022globem}, and ETRI Lifelog~\cite{oh2025understanding}. 
Among them, AI4FoodDB~\cite{ai4fooddb2023,lacruz2025ai4food} is particularly well aligned with our objective because it provides month-long longitudinal, multimodal records from 100 participants, jointly capturing dietary intake, physical activity, sleep patterns, and emotional states. Despite these advances in data collection, existing resources primarily provide raw lifelog records and lack task definitions and evaluation benchmarks for health reasoning. 
Therefore, \systemname converts structured lifestyle records into a verifiable QA benchmark for long-horizon, cross-domain personal health reasoning.

%Existing datasets cover diverse signals, including sleep, physical activity, stress, emotion, diet, and multimodal lifelog records. 
%Hence, they cannot systematically assess whether models can reason over heterogeneous records within long time windows and produce actionable health insights.

\paragraph{\textbf{Healthcare QA Benchmarks and Agents.}}
Recently, QA benchmarks for health and lifestyle have focused on some important aspects of personal health, as summarized in Table~\ref{tab:Related work comparison}.
SleepQA~\cite{bojic2022sleepqa} focuses on sleep guidance, NGQA~\cite{zhang2024ngqa} studies personalized reasoning between food, nutrition, and health states, and MentalChat16K~\cite{xu2025mentalchat16k} targets emotional and mental-health conversations.
SensorQA~\cite{reichman2025sensorqa} and OpenLifelogQA~\cite{tran2025openlifelogqa} further established benchmarks for answering daily life questions based on sensor signals and personal life logs.
Meanwhile, a series of LLM-based systems and agents for personal health have also been proposed~\citep{khasentino2025personal,fang2024physiollm,kim2024health,yu2025sensorchat,tian2025dailyllm,xu2024mental,yang2024mentallama}.
These studies indicate that LLMs have great potential in personal health analysis, but existing benchmarks and model studies still largely focus on single health dimensions or task-specific applications. A unified and verifiable benchmark remains lacking for systematically evaluating the capabilities and limitations of LLMs in long-horizon, cross-domain lifestyle health reasoning. To address this gap, \systemname introduces a QA benchmark grounded in multidimensional lifestyle records, enabling systematic evaluation of complex reasoning over long temporal horizons, heterogeneous health dimensions, and multiple users. It provides a standardized and extensible testbed, along with a strong baseline, for assessing and advancing LLMs’ personal health reasoning capabilities.

%However, real personal health reasoning often requires the simultaneous correlating of multi-domain lifestyle data, and conducting comprehensive analysis and reasoning the potential relationships among different health dimensions within a continuous time window.

\section{Dataset}
\subsection{Overview}
\label{sec:overview}
We build \systemname on AI4FoodDB~\cite{ai4fooddb2023}, a multi-modal lifestyle dataset collected from 100 participants that jointly records rich lifestyle and health-related signals.
We focus on four domains: diet, sleep, physical activity, and emotion, as they are widely recognized lifestyle factors associated with chronic disease risk.
We align the source records by user IDs and timestamps, and construct QA pairs whose answers are derived from executable queries, enabling verifiable reasoning over long time windows, cross-domain relations, and complex multi-user comparisons.
Collectively, \systemname contains 22,573 QA pairs, including 13,452 single-user queries and 9,121 multi-user queries.

\subsection{Task Taxonomy}
\label{sec:tasks}

\systemname is designed to evaluate personal health reasoning from basic fact retrieval to complex long-horizon analysis.
We organize the benchmark along three axes: reasoning type, reasoning scope, and answer complexity.

For the reasoning type, \systemname covers five categories.
\noindent (1) \textbf{\FQ} focuses on retrieving atomic facts at specific times, providing the basis for reconstructing individual behavior trajectories.
\noindent (2) \textbf{\AS} extends retrieval to longitudinal windows and requires statistical aggregation to characterize long-term behavior patterns.
\noindent (3) \textbf{\NC} computes relative differences across time periods, lifestyle domains, or individuals, revealing behavioral contrasts.
\noindent (4) \textbf{\CQ} introduces conditions such as personalized thresholds or multi-user statistics to identify anomalies and potential risks.
\noindent (5) \textbf{\TA} captures temporal dynamics, aiming to uncover emerging issues or signs of sustained improvement.
Together, these categories connect structured lifestyle records with health-related insights through rich and complex reasoning.

For the reasoning scope, questions are expanded along two dimensions: domain scope and user scope.
The domain scope ranges from analyzing a single lifestyle domain to comprehensive reasoning across multiple domains, 
and the user scope ranges from the analysis of individual users to the aggregation and comparison across multiple users. 
For the answer complexity, \systemname includes Yes/No labels, scalar numerical values, short text spans, pairwise outputs, and multi-item lists.
These challenging tasks allow \systemname to evaluate the capabilities and key bottlenecks of LLM agents in comprehensive health reasoning, including evidence retrieval, long-window aggregation, cross-domain alignment, multi-user comparison, and structured answer synthesis.

%%%%%%%%%%%%%
\subsection{Dataset Generation}
\label{sec:pipeline}

We design an automated and extensible pipeline to generate the QA pairs in \systemnamenospace.
First, we define a compositional query space over structured lifestyle records, and write executable programs with verifiable factual answers for each type of natural-language question.
We represent the life records of user $u$ as $\mathcal{X}_u = \{X_u^{D}, X_u^{S}, X_u^{A}, X_u^{E}\}$, where $D$, $S$, $A$, and $E$ denote the diet, sleep, physical activity, and emotion domains, respectively.
Each $X_u^{\cdot}$ is a time-stamped sequence in its corresponding domain.
All data are aligned by user IDs and timestamps, enabling queries over specific time windows and allowing evidence to be combined across domains.

We derive answers by composing a set of reasoning operators, including time-window selection, temporal alignment across domains, aggregation (e.g., mean, min, and max), comparison across time periods or user groups, conditional filtering with thresholds, and temporal pattern analysis such as consecutiveness and trends.
By composing these operators within and across domains, we instantiate queries ranging from single-domain retrieval to long-window, cross-domain, and multi-user reasoning.
For example, a query may analyze whether decreases in deep sleep duration over several consecutive days are accompanied by increases in stress levels, which requires joint reasoning over sleep and emotion records within synchronized time windows.
Our operator-based design covers all reasoning types in Sec.~\ref{sec:tasks}, while keeping each question interpretable and executable.

For each instantiated query type, we define an executable program $\pi(Q)$ to generate a natural-language question $Q$ and obtain a deterministic answer from the database, enabling verifiable evaluation of LLM responses.
We further implement programmatic checks for all QA pairs and conduct manual inspection to ensure data quality.
The pipeline is scalable and publicly released, and can be extended to new data sources or lifestyle domains by adding database schemas and query templates.
\textbf{\textit{Examples.}} Several example queries are as follows:
\textit{(i) Single-dimensional:} ``For each day last week, what was my average sleep duration?''
\textit{(ii) Cross-dimensional:} ``Over the past week, on how many days did I exceed both an activity-duration [threshold] and a sleep-duration [threshold]?''
\textit{(iii) Multi-user:} ``Over the past week, on how many days did my sleep duration and aerobic activity time both exceed the cohort average? On those days, what was my dominant diet category, and what trend did my emotion score exhibit?''

%%%%%%%%%%%%%%
%\item \textit{Multi-user:} ``Do the lowest-stress users have higher average activity and sleep duration than others?''
%Instantiated by ranking users by average stress level, selecting the bottom group (e.g., bottom 10\%), and comparing group means for activity and sleep.
%\end{itemize}

%%%%%%%%%%%

\begin{figure*}[t]
\begin{center}
    \includegraphics[width=0.8\linewidth]{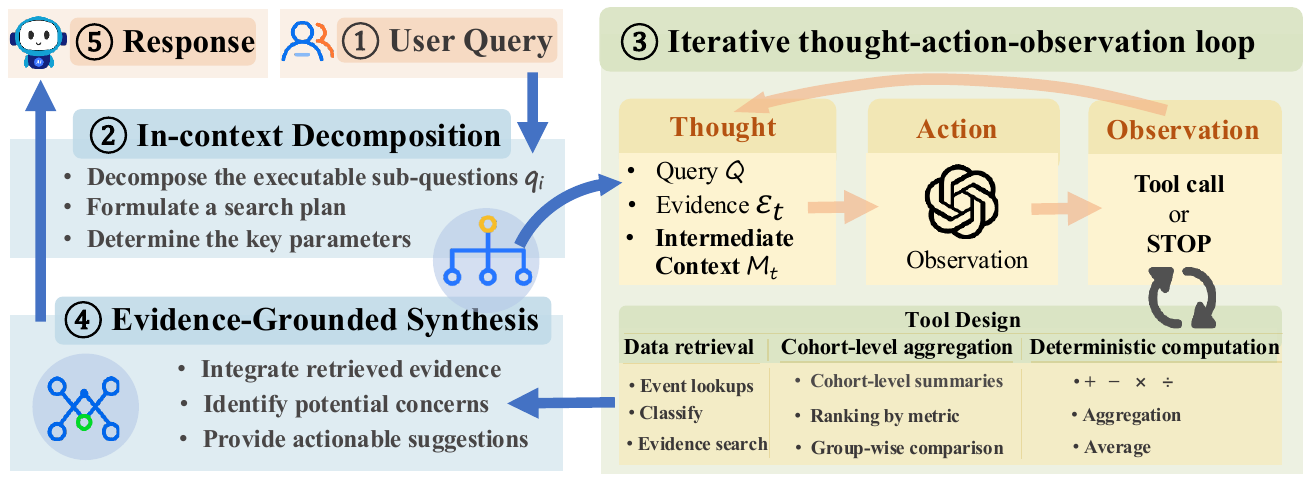}
\end{center}
\vspace{-0.4cm}
\caption{Overview of the \agentmname framework. \agentmname decomposes user queries, retrieves evidence through iterative tool use, and synthesizes evidence-grounded responses.}
\label{fig:agentframework}
\vspace{-0.3cm}
\end{figure*}

\section{LifeAgent}
\label{sec:lifeagent}
%Our extensive evaluation on \systemname shows that current LLMs still face clear bottlenecks in complex long-horizon and cross-domain lifestyle reasoning.
%We therefore propose \systemnamenospace, a training-free health agent that performs multi-step reasoning through tool calls to identify the evidence required by a query, derive intermediate results, and synthesize the final answer.

\subsection{Motivation}

Our evaluations on \systemname (Section~\ref{sec:results}) show that LLMs struggle with complex health reasoning tasks, especially long-window aggregation and cross-dimensional evidence integration.
These bottlenecks suggest that reliable lifestyle health reasoning requires more than well-designed prompting: the system must identify relevant evidence, decompose the task into executable steps, and perform aggregation and comparison based on explicitly retrieved evidence and rigorous computation. This observation motivates our tool-augmented design, which separates high-level reasoning from evidence retrieval and deterministic computation.
%We design \agentmname as a training-free reasoning baseline that performs multi-step reasoning by decomposing complex health reasoning tasks into sub-questions, iteratively retrieving relevant evidence, and invoking tools for deterministic aggregation and comparison.
%This design allows the LLM to focus on high-level planning and response synthesis, while tools handle grounding-intensive retrieval and computation.

\subsection{Agent Workflow}
\label{sec:agent_workflow}

Following recent agentic reasoning frameworks~\cite{smolagents}, we structure \agentmname around a thought-action-observation loop, as illustrated in Figure~\ref{fig:agentframework}. Given a user-facing health query $Q$, \agentmname aims to generate an evidence-grounded response $R$ by retrieving and analyzing relevant lifestyle records.
For each complex query $Q$, \agentmname first decomposes it into a set of sub-questions $\{q_i\}_{i=1}^{k}$ and constructs an executable plan $P$.
At step $t$, the agent maintains an evidence cache $\mathcal{E}_t$, which stores records retrieved through tool calls, and an intermediate context $\mathcal{M}_t$, which stores derived results (e.g., aggregates, comparisons, detected anomalies, and trend summaries).
When new observations become available, the agent updates the plan by adding or revising sub-questions, and continues to invoke tools to incrementally update $\mathcal{E}$ and $\mathcal{M}$.

This multi-step decomposition is evidence-driven and adaptive, which improves the ability of LLMs to analyze complex queries.
For example, for a comprehensive query such as ``How was my sleep quality last week?'', \agentmname first retrieves daily sleep-related indicators within the time window, such as sleep duration, summarizes patterns, and identifies abnormal days.
When potential anomalies are detected, it further drills down into event-level records on those days, such as irregular diet or unusual activity, and stores the findings as supporting evidence.
After all sub-questions in the current plan are executed, \agentmname synthesizes the final answer conditioned on the original query $Q$, the accumulated evidence $\mathcal{E}$, and the intermediate context $\mathcal{M}$:
$R = \mathrm{Synthesize}(Q, \mathcal{E}, \mathcal{M})$.
This workflow enables \agentmname to answer complex health queries through explicit evidence retrieval, adaptive decomposition, deterministic computation, and evidence-grounded synthesis.

\subsection{Tool Design}
\label{sec:tool_design}

To support multi-step decomposition and reliable iterative analysis in \agentmnamenospace, we design three categories of tools.
These tools allow the agent to iteratively retrieve evidence, derive intermediate statistics, and provide reliable computational support for complex reasoning and decision-making.

\textbf{(1) Structured data retrieval.} These tools provide structured access to the lifestyle-record database under explicit constraints, including domain, time window, and granularity.
They support event-level lookups and daily aggregated time-series extraction.
\textbf{(2) Cohort-level aggregation.}
Multi-user queries require aggregating statistics across users. We design general cohort-level operators to compute multi-user summaries (e.g., mean, median, percentiles), rank users by selected metrics, and split users into groups for criterion-based comparisons, enabling efficient multi-user aggregation and comparison.
\textbf{(3) Deterministic computation.}
These tools deterministically perform basic arithmetic, aggregation, threshold checking, comparison, and temporal pattern analysis, such as consecutiveness and trends.
By delegating low-level numerical and aggregation operations to deterministic tools, \agentmname reduces computation-induced LLM hallucinations and separates reliable calculation from high-level reasoning, allowing the LLM to focus on query decomposition, reasoning control, and evidence-grounded synthesis.

%\section{Experiments}\label{sec:experiment}
%We define two evaluation settings and perform a comprehensive evaluation of eight widely used open-source LLMs and three closed-source LLMs on \systemnamenospace. We first compute and compare the overall accuracy of all models across the complete set of reasoning questions. Subsequently, we perform a detailed analysis along multiple dimensions, including comparisons by question and answer distribution, as well as differences across dimensions and user settings.
%\red{\textbf{XXXX, will revise later}}
% We further investigate how model performance changes as the temporal range of reasoning data expands, and assess the impact of quantization strategies on both accuracy and inference latency.

%%%%%%%%%%%%%%%%%%%%%%%%%%%%%%%%%%%%%
\begin{table}[t]
\centering
\resizebox{0.88\linewidth}{!}{%
\begin{tabular}{l|c|cccc}
\toprule
Model & \multicolumn{1}{c|}{CP} & \multicolumn{4}{c}{DP} \\
\midrule
Metrics & Acc (\%) & Acc (\%) & VA(\%) & EX(\%) & Acc$\mid$EX(\%) \\
\midrule
\multicolumn{6}{c}{\textbf{Open-source LLMs}} \\
\midrule
\rowcolor{gray!15}Deepseek-Coder-1.3B & 1.09 & 1.26 & 43.83 & 14.62 & 4.00 \\
Llama-3.2-3B        & 20.18 & 13.47 & 47.29 & 17.99 & 67.68 \\
\rowcolor{gray!15}Phi-3.5-mini-3.8B   & 20.57 & 16.16 & 56.67 & 20.23 & 77.34 \\
Mistral-v0.3-7B     & 30.97 & 9.03 & 28.46 & 11.48 & 75.35 \\
\rowcolor{gray!15}Qwen-2.5-7B         & 40.45 & 21.45 & 55.83 & 24.71 & 84.78 \\
Llama-3.1-8B        & 20.65 & 21.53 & 63.33 & 27.25 & 77.73 \\
\rowcolor{gray!15}Gemma-2-9B          & 24.44 & 14.54 & 56.24 & 27.85 & 51.15 \\
Qwen3.5-9B          & 45.55 & 25.76 & 63.03 & 32.81 & 78.51 \\
\rowcolor{gray!15}Llama-3.1-70B       & 40.51 & 13.91 & 45.77 & 22.73 & 58.41 \\
\midrule
\multicolumn{6}{c}{\textbf{Closed-source LLMs}} \\
\midrule
\rowcolor{gray!15}Gemini 2.5 Flash-Lite     & 44.81 & \underline{39.04} & \underline{84.84} & \underline{45.97} & 82.92 \\
Claude-3-Haiku      & 35.30 & 29.30 & 74.06 & 36.49 & 75.71 \\
\rowcolor{gray!15}GPT-4o              & \underline{57.02} & 34.71 & 63.85 & 35.97 & \textbf{95.65} \\
DeepSeek-V4-Pro     & \textbf{58.89} & \textbf{55.67} & \textbf{89.45} & \textbf{56.15} & \underline{90.81} \\
\bottomrule
\end{tabular}}
\vspace{-0.1cm}
\captionof{table}{Overall results of LLMs on \systemnamenospace.}
\label{tab:overallresults}
\vspace{-0.3cm}
\end{table}
%%%%%%%%%%%%%%%%%%%%%%%%%%%%%%%%%%%%%

\section{Experimental Setup}
\textbf{Models.} 
We evaluate a diverse set of representative LLMs, covering both open-source and closed-source models with a wide range of scales.
\textit{(1) Open-source models}: DeepSeek-Coder-1.3B~\cite{guo2024deepseek}, Llama~\cite{dubey2024llama} (Llama-3.2-3B; Llama-3.1-8B/70B), 
Phi-3.5-mini-3.8B~\cite{abdin2024phi3technicalreporthighly},
Mistral-v0.3-7B~\cite{jiang2023mistral7b},
Qwen-2.5-7B~\cite{qwen2.5},  Qwen3.5-9B~\cite{team2026qwen3} and Gemma-2-9B~\cite{team2024gemma}.
\textit{(2) Closed-source models}: GPT-4o~\cite{openai2024gpt4o}, Gemini 2.5 Flash-Lite~\cite{comanici2025gemini}, Claude-3-Haiku~\cite{anthropic2024claude3} and DeepSeek-V4-Pro~\cite{xu2026deepseek}.

\textbf{Baselines.} 
We evaluate LLMs on \systemname under two baseline settings: 
\textit{(1) Context Prompting (CP)~\cite{sui2024table}}.
Given a health reasoning question, the system first pre-filters the life record data to keep the target user’s data, and then embeds the filtered data with the question into the prompt for the LLM to reason.
\textit{(2) Database-augmented Prompting (DP)~\cite{biswal2024text2sql}}. 
DP adopts a two-stage, LLM-driven retrieval-then-reasoning workflow.
Given the question and the database schema, the LLM generates an SQL query to retrieve evidence from the database.
The system validates the query and executes only complete read-only SQL queries; the returned results are then fed back to the LLM to produce the final answer.

\textbf{Metrics.}
We use \emph{Accuracy (Acc)} as the primary metric across all settings:
a prediction is correct if it matches the normalized ground truth.
Yes/No and text answers require exact match; numeric answers allow an
error of $\pm 1$ when $gt \le 14$ and
$\max(0.5\%\cdot |gt|, 0.01)$ otherwise; pairwise and multi-item
answers must contain the same number of items, with each item correctly
matched.
For DP, we further report three automatically computed metrics.
For instance $i$, let $s_i$ be the generated SQL, $E_i$ its returned
evidence, $E_i^\ast$ the evidence required for the complete ground-truth
answer, and $y_i$ and $\hat{y}_i$ the ground-truth and predicted answers.
Let
$\mathrm{VA}_i=\mathbf{1}[s_i\text{ executes successfully}]$ and
$\mathrm{EX}_i=\mathbf{1}[\mathrm{VA}_i=1 \land E_i^\ast\subseteq E_i]$.
We define
\begin{flalign*}
&\mathrm{VA}
= \frac{1}{N}\sum_{i=1}^{N}\mathrm{VA}_i, &&\\[2pt]
&\mathrm{EX}
= \frac{\sum_i \mathrm{EX}_i}{\sum_i \mathrm{VA}_i}, &&\\[2pt]
&\mathrm{Acc}\mid\mathrm{EX}
= \frac{
\sum_i \mathbf{1}[\mathrm{EX}_i=1 \land \hat{y}_i=y_i]
}{
\sum_i \mathrm{EX}_i
}. &&
\end{flalign*}
Here, $\hat{y}_i=y_i$ denotes a correct match under the normalization
rules above. Partial evidence is counted as incorrect.
Thus, $\mathrm{Acc}\mid\mathrm{EX}$ measures final-answer accuracy
conditioned on successful evidence retrieval.

%Benchmarking 
\section{Results}\label{sec:results}
%\textbf{Overall Results.}
We organize our experiments around four research questions:
(1) How well do existing LLMs perform on \systemnamenospace?
(2) Where do LLMs fail as reasoning complexity increases?
(3) Can \agentmname enhance LLMs' reasoning capabilities?
(4) How does each component contribute to \agentmnamenospace?

%%%%%%%%%%%%%%%%%%%
\begin{figure}[t]
\centering
\includegraphics[width=0.8\linewidth]{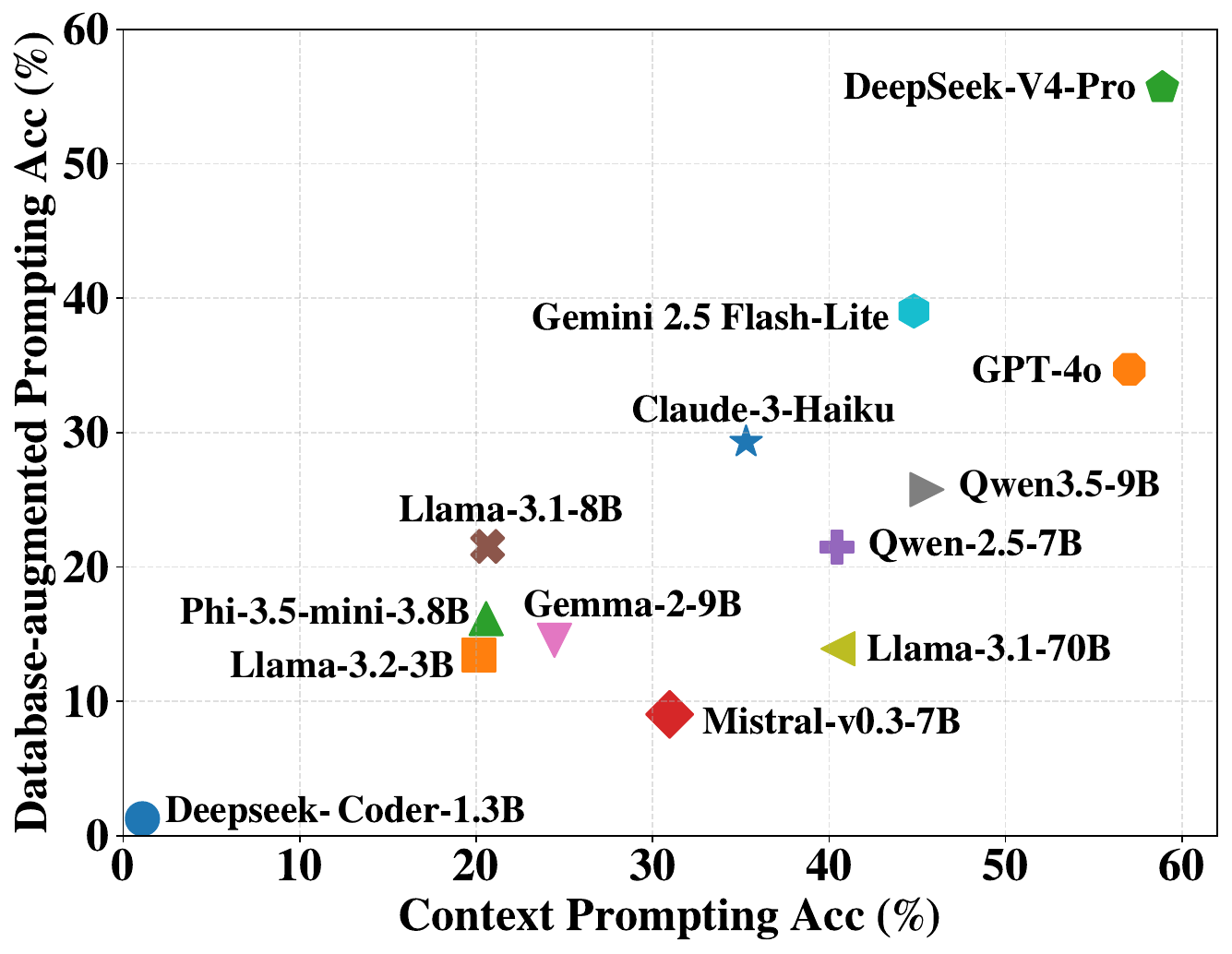}
\vspace{-0.3cm}
\caption{Accuracy (\%) comparison of all LLMs under two evaluation settings: CP vs. DP.}
\label{fig:overall}
\vspace{-0.3cm}
\end{figure}
%%%%%%%%%%%%%%

\paragraph{\textbf{Question 1: How well do existing LLMs perform on \systemnamenospace?}}
Table~\ref{tab:overallresults} and Figure~\ref{fig:overall} summarize the performance of all evaluated LLMs on the full dataset.
Closed-source models overall outperform open-source models, and the recent models show improved overall performance.
Under CP, DeepSeek-V4-Pro achieves the highest accuracy of 58.89\%, while Qwen3.5-9B is the best-performing open-source model with 45.55\%.
Under DP, DeepSeek-V4-Pro again performs best with 55.67\%, followed by Gemini 2.5 Flash-Lite with 39.04\%; among open-source models, Qwen3.5-9B achieves the highest accuracy of 25.76\%.
These results reveal a clear capability gap of current LLMs in performing long-horizon health reasoning over large-scale, cross-dimensional life records.
However, we find that the final-answer accuracy improves substantially when complete evidence is successfully retrieved: nine models achieve Acc$\mid$EX above 70\%, and GPT-4o reaches 95.65\%.
In contrast, EX remains much lower, averaging only 28.79\% across models.
Even for DeepSeek-V4-Pro, EX reaches only 56.15\%.
These results indicate that reliable evidence retrieval remains a major bottleneck even for recent models.

\noindent\textbf{Finding 1.}
Despite recent model improvements, existing LLMs still struggle on \systemname, with reliable evidence retrieval being a key bottleneck.
Moreover, current models remain limited in generating effective database queries and reliably retrieving evidence.

\begin{figure}[t]
\centering
\includegraphics[width=1\linewidth]{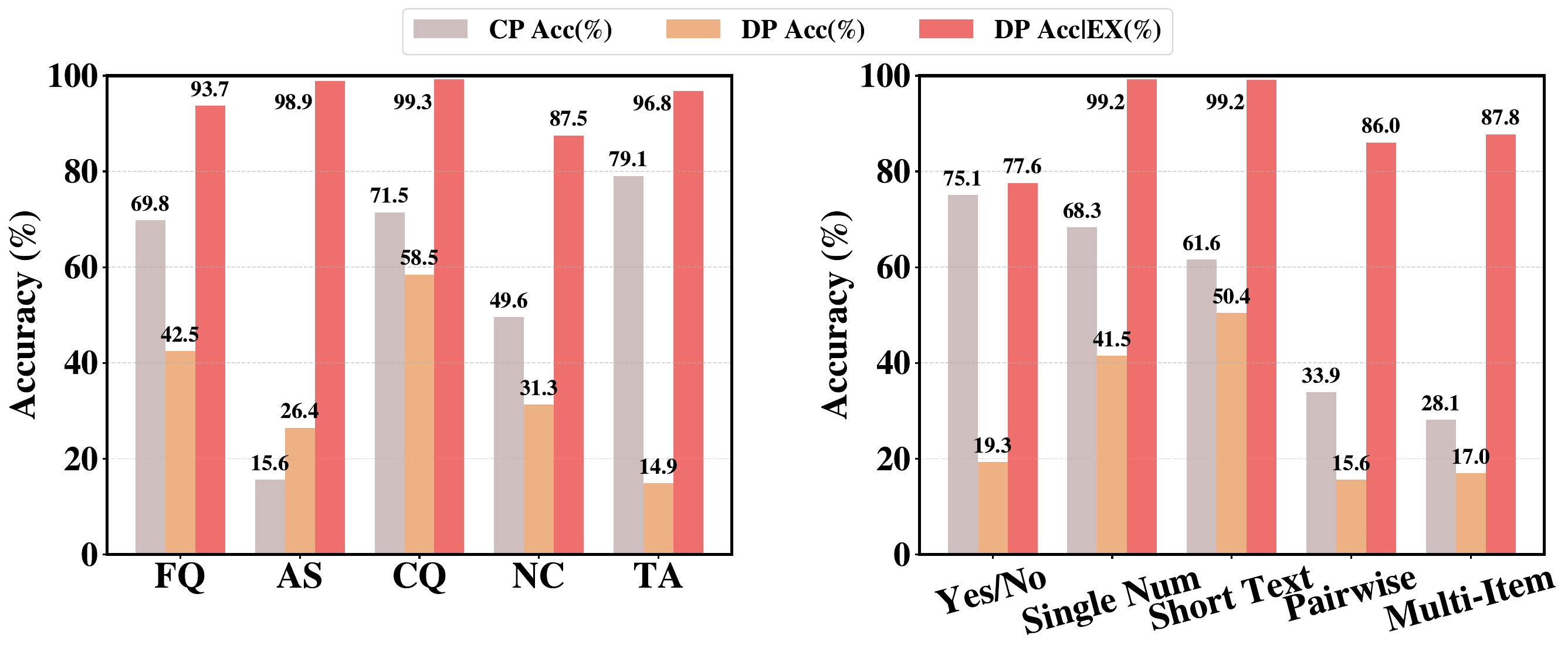}
\vspace{-0.5cm}
\caption{
GPT-4o performance across question types (left) and answer formats (right) in CP and DP.}
\label{fig:results_UandD}
\vspace{-0.4cm}
\end{figure}
%Average accuracy of all models across different dimensions and user settings \textbf{(right)}

\paragraph{\textbf{Question 2: Where do LLMs fail as reasoning complexity increases?}}
To further analyze the bottlenecks of current LLMs, we analyze their reasoning performance across different task types, answer formats, evidence dimensions, and user settings.
First, models degrade substantially on aggregation-intensive tasks and multi-item outputs.
By task type, models perform relatively well on fact-oriented queries such as FQ, but struggle with more complex reasoning.
For AS, the average accuracy across models included in our fine-grained analysis is only 5.72\% under CP and 15.31\% under DP, with NC showing similarly low performance.
By answer type, simple outputs such as Yes/No and single-number answers are handled relatively well, while multi-item  outputs remain difficult.
For multi-item answers, the average accuracy drops to 8.62\% under CP and 3.54\% under DP.
These difficulties persist even for strong proprietary models: as shown in Figure~\ref{fig:results_UandD}, GPT-4o performs well on easier categories but still struggles with AS questions and complex outputs.

Second, performance drops as the evidence scope expands across dimensions and users.
Across dimensions, the average accuracy in the fine-grained analysis decreases from 41.54\% to 23.42\% under CP and from 30.84\% to 9.63\% under DP when moving from single-dimension to all-dimension tasks.
Multi-user settings further increase the difficulty, as they require cohort-level aggregation and comparison across users.
Under the CP setting, the required data can easily exceed the LLM context-window limit, directly leading to reasoning failure. Under DP, LLM performance also drops substantially, indicating that current LLMs still have major limitations in large-scale evidence retrieval and aggregation.

\noindent\textbf{Finding 2.}
LLMs struggle most with aggregation-intensive reasoning and structured outputs, and the performance further degrades as the evidence scope expands across more lifestyle dimensions and users.

%%%%%%%%%%%%%%%%%%%%%%%%%%%%%%%%%%%%%%%%%
\begin{table}[t]
\centering
\newcolumntype{K}{wc{1.5cm}} 
\resizebox{0.9\linewidth}{!}{%
\begin{tabular}{l | l | K K K}
\toprule
\multicolumn{2}{c}{\textbf{The most difficult category}}  & 
\textbf{CP} & \textbf{DP} & \textbf{LifeAgent} \\
\midrule
% ---------------- Question Type ----------------
\multirow{2}{*}{\textbf{Question Type}} 
& \cellcolor{gray!15}AS & \cellcolor{gray!15}5.21 & \cellcolor{gray!15}17.35 & \cellcolor{gray!15}24.71 \\
& NC & 24.42 & 12.29 & 40.15 \\
\midrule
% ---------------- Answer Type ----------------
\multirow{2}{*}{\textbf{Answer Type}} 
& \cellcolor{gray!15}Pairwise Answer & \cellcolor{gray!15}1.78 & \cellcolor{gray!15}6.06 & \cellcolor{gray!15}38.31 \\
& Multi-item Answer & 0.36 & 3.04 & 32.31 \\
\midrule
%--------------- Cross-dimension ----------------
\multirow{2}{*}{\textbf{All dimensions}} 
& \cellcolor{gray!15}Single-user & \cellcolor{gray!15}14.69 & \cellcolor{gray!15}2.81 & \cellcolor{gray!15}53.06 \\
% ---------------- Multi-user ----------------
& Multi-user & 0.00 & 15.00 & 52.44 \\
\midrule
\textbf{Average}& \cellcolor{gray!15}-- & \cellcolor{gray!15}\textbf{7.74} & \cellcolor{gray!15}\textbf{9.43} & \cellcolor{gray!15}\textbf{40.16} \\
\bottomrule
\end{tabular}%
}
\vspace{-0.1cm}
    \caption{Accuracy (\%) on the most challenging subsets of \systemnamenospace, comparing \agentmname~with CP and DP baselines under the same backbone (Qwen2.5-7B).}
\label{tab:lifeagentresults}
\vspace{-0.4cm}
\end{table}
%%%%%%%%%%%%%%%%%%%%%%%%%%%%%%%%%%%%%%%%%
%(\small N/A is because the CP baseline cannot complete multi-user reasoning.)

\paragraph{\textbf{Question 3: Can \agentmname enhance LLMs' reasoning capabilities?}}

Previous analyses show that current LLMs have clear limitations in aggregation-intensive reasoning, structured answer synthesis, and large-scope evidence retrieval.
To enhance LLMs' reasoning capabilities in these challenging cases, \agentmname decomposes complex queries into multi-step reasoning processes and uses a suite of tools for evidence retrieval and computation, combining tool-based computation with the high-level reasoning ability of LLMs to support cross-domain and multi-user reasoning.
We compare \agentmname with baselines on the most challenging reasoning subsets, as shown in Table~\ref{tab:lifeagentresults}.
Overall, \agentmname substantially improves accuracy on these tasks: the average accuracy increases from 7.74\% with CP and 9.43\% with DP to 40.16\%, corresponding to gains of +32.42 and +30.73 points, respectively.
Specifically, \agentmname markedly improves performance on the hardest question types, such as AS and NC, indicating stronger capability for aggregation-intensive reasoning.
It also improves compositional-output reasoning: on multi-item answers, CP and DP achieve only 0.36\% and 3.04\% accuracy, whereas \agentmname reaches 32.31\%.
Moreover, \agentmname scales better to larger evidence scopes, raising all-dimension accuracy to above 50\% for both single-user and multi-user queries, while CP and DP are below 15\%.

\noindent\textbf{Finding 3.}
\agentmname enhances LLMs' reasoning capabilities on challenging subsets, especially for aggregation-intensive reasoning, structured answer synthesis, and large-scope cross-domain and multi-user evidence integration.

%%%%%%%%%%%%%%%%%%%%%%%%%%%%%%%%%%%%%%%%%
\begin{table}[t]
\centering
\newcolumntype{K}{wc{2.7cm}}
\resizebox{\linewidth}{!}{%
\begin{tabular}{l | K K}
\toprule
\textbf{} &
\makecell{\textbf{All-Dim. }\\\textbf{Benchmark(\%)}} &
\makecell{\textbf{Health-Indicator}\\\textbf{Queries (\%)}} \\
\midrule
\rowcolor{gray!15}
w/o Decomposition Loop & 38.76 & 61.97 \\
w/o Retrieval Tools & 46.13 & 33.84 \\
\rowcolor{gray!15}
w/o Computation Tools & 43.65 & 66.37 \\
\midrule
\textbf{Full \agentmname} & \textbf{52.75} & \textbf{69.84} \\
\bottomrule
\end{tabular}%
}
\vspace{-0.1cm}
\caption{Ablation study of \agentmnamenospace. Accuracy (\%) is reported on all-dimension benchmark questions and health-indicator queries.}
\label{tab:ablation}
\vspace{-0.4cm}
\end{table}
%%%%%%%%%%%%%%%%%%%%%%%%%%%%%%%%%%%%%%%%%

\paragraph{\textbf{Question 4: How does each component contribute to \agentmnamenospace?}}
To study the contributions of \agentmnamenospace's key components, we conduct an ablation study.
Since rerunning all ablation variants over the full 22,573-question benchmark is computationally expensive, we evaluate them on the all-dimension benchmark questions and the health-indicator queries from the case study described in next section.
These two evaluation sets provide complementary coverage of challenging cross-dimensional reasoning and detailed lifestyle analysis.
All variants use the same Qwen2.5-7B backbone, question sets, decoding settings, and evaluation protocol, with only the corresponding component removed.
As shown in Table~\ref{tab:ablation}, the full \agentmname performs best on both evaluation sets, and removing any component consistently degrades performance.
On the all-dimension benchmark questions, removing decomposition causes the largest drop (52.75\% to 38.76\%), highlighting its importance in avoiding omitted cross-domain sub-tasks and incomplete answers.
On the health-indicator queries, removing retrieval tools has the greatest impact (69.84\% to 33.84\%), highlighting the importance of complete and accurate evidence acquisition.
Removing the computation tools also degrades performance, supporting their role in reducing errors in long-window aggregation and comparison.

\noindent\textbf{Finding 4.}
The ablation study confirms that the three key components of \agentmname make complementary contributions: decomposition is particularly important for complex cross-dimensional reasoning, retrieval is critical for complete evidence acquisition, and deterministic computation supports reliable long-window aggregation and comparison.

\section{Case Study: Personal Health Assistants}\label{sec:case_study}
\paragraph{\textbf{Question 5: Can \agentmname support realistic personal health assistant scenarios?}}
\agentmname demonstrates the feasibility of serving as a strong baseline for reliable health inference on lifestyle data.
In realistic usage scenarios, users expect end-to-end assistance for health queries and reasoning, including evidence-grounded retrieval, cross-domain integration, and actionable, constructive suggestions based on their lifestyle records.
To evaluate the potential of \agentmname as a personal health assistant, we conduct a case study grounded in everyday usage scenarios, covering three realistic task types:
(1) \textit{Health-indicator queries}: factual status retrieval, e.g., ``How was my [indicator(s)] this week?'', where indicators may come from one or multiple domains, such as sedentary duration and sleep duration.
(2) \textit{Comprehensive health reasoning}: holistic status assessment and issue identification over multiple domains, e.g., ``How was my overall condition this week, and are there any issues I should be aware of?'', integrating activity, sleep, emotion, and diet.
(3) \textit{Targeted lifestyle recommendations}: actionable suggestions grounded in the assessed issues and supporting evidence, e.g., ``Based on my past-week condition, what targeted changes should I make?''.

%%%%%%%%%%%%%%%%%%%%%%%%%%%%%%%%%%%%%%%%%%%%%
\begin{table*}[t]
\centering
\newcolumntype{K}{wc{1.8cm}} 
\resizebox{\linewidth}{!}{%
\begin{tabular}{l | K | K | K | K | K | K | K | K | K }
\toprule
% ---- Header row 1 ----
\rowcolor{white}
\textbf{Task}
& \multicolumn{3}{c}{\textbf{Health-indicator queries (Acc \%)}}
& \multicolumn{3}{c}{\textbf{Comprehensive health reasoning (Score 1--5)}}
& \multicolumn{3}{c}{\textbf{Targeted lifestyle recommendations (Score 1--5)}} \\
\cmidrule(lr){2-10}
% ---- Header row 2 ----
\rowcolor{white}
\textbf{Dimension}
& \textbf{CP} & \textbf{DP} & \textbf{LifeAgent}
& \textbf{CP} & \textbf{DP} & \textbf{LifeAgent}
& \textbf{CP} & \textbf{DP} & \textbf{LifeAgent} \\
\midrule
% ---- Data rows ----
\rowcolor{gray!15}
Activity
& 40.22 & 59.40 & 78.56
& 1.63 & 2.08 & 3.16
& 2.62 & 2.84 & 3.09 \\

Sleep
& 38.86 & 39.62 & 71.51
& 2.01 & 2.11 & 3.05
& 3.26 & 3.56 & 3.77 \\

\rowcolor{gray!15}
Emotion
& 49.70 & 60.47 & 79.68
& 1.89 & 2.34 & 3.40
& 2.69 & 2.91 & 3.24 \\

Diet
& 27.27 & 17.14 & 49.34
& 1.32 & 1.42 & 3.03
& 2.60 & 2.76 & 3.12 \\

\rowcolor{gray!15}
All dimensions
& 36.21 & 40.15 & 70.12
& 2.14 & 2.46 & 2.70
& 2.51 & 2.55 & 2.87 \\

\midrule
\textbf{Average}
& \textbf{38.45} & \textbf{43.36} & \textbf{69.84}
& \textbf{1.80} & \textbf{2.08} & \textbf{3.07}
& \textbf{2.74} & \textbf{2.92} & \textbf{3.22} \\
\bottomrule
\end{tabular}%
}
\caption{Case-study results of CP, DP, and \agentmname on lifestyle health reasoning tasks.
Acc (\%) is reported for health-indicator queries, while rubric-based scores averaged across five LLM judges are reported for comprehensive reasoning and targeted recommendations.}
\label{tab:casestudy}
\vspace{-0.4cm}
\end{table*}
%%%%%%%%%%%%%%%%%%%%%%%%%%%%%%%%%%%%%%%%%%%%%

\textbf{Setup and Metrics.}
We conduct the case study using the 100 users in \systemnamenospace,
without collecting any new user data.
For each user, we construct 10 health-indicator queries,
5 comprehensive assessment queries, and 5 recommendation queries.
We evaluate at two levels: for Task (1), where answers can be deterministically verified against the underlying records, we report Accuracy.
For Tasks (2) and (3), where no standard ground truth exists, we follow
recent LLM-as-a-judge evaluation practices for open-ended generation and
instruction following~\cite{zheng2023judging,liu2023g,li2025generation,huang2025empirical}.
Specifically, we use five LLM judges: Gemini-3.5-Flash~\cite{gemini35flash},
DeepSeek-V4-Pro~\cite{xu2026deepseek}, Claude-Sonnet-5~\cite{claudesonnet5}, Qwen3.5-9B~\cite{team2026qwen3}, and GPT-4o~\cite{openai2024gpt4o}.
All judges are provided with the same user task, supporting evidence as
the source of truth, candidate response, and evaluation rubric, and assign
1--5 scores on faithfulness, aggregation correctness, coverage,
actionability, personalization, and conciseness.
We report scores averaged across the five judges.

\noindent\textbf{Results and Discussion.}
Table~\ref{tab:casestudy} reports the performance of \agentmname on the three assistant-style tasks, compared with the two baselines.
The results show that \agentmname achieves higher performance than both CP and DP across all three task types.
For health-indicator queries, \agentmname obtains 69.84\% average accuracy, outperforming CP (38.45\%) and DP (43.36\%).
For the two open-ended tasks, the five-judge average scores consistently favor \agentmnamenospace.
On comprehensive health reasoning, \agentmname scores 3.07, compared with 1.80 for CP and 2.08 for DP; on targeted lifestyle recommendations, it scores 3.22, compared with 2.74 and 2.92, respectively.
Moreover, \agentmname achieves the highest average score under each individual judge on both open-ended tasks, indicating that the case-study conclusions are consistent across different LLM judges rather than being driven by a single judge.

These gains suggest that multi-step reasoning and tool design enable \agentmname to strengthen LLMs for complex cross-domain lifestyle reasoning, allowing it to better support practical end-to-end health assistant scenarios.
Importantly, these improvements are built on \systemnamenospace: evaluations on \systemname reveal key failure modes of current LLMs in long-horizon, cross-domain lifestyle reasoning, which motivate our analysis and the design of \agentmnamenospace.
Therefore, \systemname can serve as a diagnostic testbed for future work.
As new models and agents are evaluated, newly revealed bottlenecks can in turn guide improved model designs, reasoning strategies, and tool-use mechanisms for grounded long-horizon reasoning.

\section{Conclusion}
We introduce \systemnamenospace, a large-scale benchmark for evaluating LLMs on long-horizon, cross-dimensional, and multi-user lifestyle health reasoning over personal records.
Using \systemnamenospace, we systematically evaluate 13 representative LLMs and uncover clear bottlenecks: while models perform relatively well on simple factual retrieval, they degrade substantially on long-horizon aggregation, cross-dimensional evidence integration, and multi-user reasoning.
Motivated by these findings, we propose \agentmnamenospace, a tool-augmented baseline that decomposes complex queries, performs multi-step evidence retrieval, and invokes tools for deterministic aggregation over long time horizons.
 \agentmname substantially improves performance on the most challenging tasks, demonstrating the value of explicit evidence grounding and deterministic tool use for lifestyle health reasoning.
We release the benchmark, construction pipeline, and evaluation protocol to support future research on LLMs and agents with stronger long-horizon, cross-dimensional reasoning capabilities.

\section*{Limitations}
Although \systemname provides a large-scale and verifiable benchmark for
long-horizon lifestyle health reasoning, its current instantiation is
constructed primarily from a specific lifestyle dataset and focuses on four
domains: diet, sleep, physical activity, and emotion.
These domains capture important aspects of daily health behavior, but do not
provide a comprehensive representation of personal health, such as clinical
measurements, medical history, medication use, or other physiological signals.
Further validation is therefore needed across additional health factors,
longer time spans, sensing modalities, populations, and data-collection
settings.
The released construction and evaluation framework can be extended to
datasets containing such additional information, providing a direction for
broader evaluation over heterogeneous health records.

Second, the QA pairs are generated through a template- and operator-based
pipeline.
This controlled design enables scalable benchmark construction and
deterministic ground-truth derivation, which are important for reliable and
reproducible evaluation.
However, it may not fully capture the diversity and complexity of natural
user interactions.
Although our case study extends the evaluation to more open-ended
life assistant-style scenarios, the current work does not systematically evaluate
ambiguity, underspecification, paraphrastic diversity, informal wording, or
multi-turn conversational context.
Extending the benchmark to these more complex language and dialogue settings
is an important direction for future work.

Finally, the current benchmark is constructed from quality-controlled records
after preprocessing missing and anomalous values.
Consequently, our experiments do not systematically evaluate the robustness
of \agentmname or the evaluated LLMs to unprocessed missing, noisy, or
unreliable records.
Future systems could incorporate data-completeness checks, anomaly detection,
and explicit handling of uncertain evidence before reasoning and synthesis.
Further validation is also needed under more realistic deployment conditions,
including latency, user interaction, and expert review.

\section*{Ethical Considerations}
\systemname is designed as a benchmark for evaluating long-horizon
lifestyle health reasoning rather than as a clinical decision-making
system.
It is constructed from AI4FoodDB~\cite{ai4fooddb2023}, an existing
dataset released for research purposes, in which personal information
has been anonymized and participants are represented by IDs.
We do not collect new human-subject data or introduce additional
personal information.
The released benchmark is intended to support research on
record-grounded reasoning over structured lifestyle data.
Since AI4FoodDB~\cite{ai4fooddb2023} was collected from a specific study
population, \systemname may inherit demographic, geographic, cultural,
and behavioral biases from the original data and may not fully represent
broader populations or health conditions.
Therefore, results on \systemname should be interpreted as evaluations
of model reasoning over this structured lifestyle dataset rather than as
direct evidence of generalization to broader real-world populations or
health contexts.
Similarly, \systemname and \agentmname are intended for benchmarking and
research on evidence-grounded lifestyle reasoning, not for diagnosis,
treatment, or medical decision-making.
Automatically generated lifestyle assessments or recommendations should not be treated as professional
medical advice or as a substitute for qualified medical judgment.
Deploying similar systems in real-world health settings would require
additional safeguards and validation, including privacy protection, user
consent, safety mechanisms, and expert review.

\section*{Acknowledgements}
This work has been funded in part by NSF, with award numbers \#2112665, \#2112167, \#2003279, \#2120019, \#2211386, \#2052809, \#1911095 and in part by PRISM and CoCoSys, centers in JUMP 2.0, an SRC program sponsored by DARPA.

\bibliography{custom}

%\section{Example Appendix}
%\label{sec:appendix}

\appendix

\begin{figure*}[t]
\centering
\includegraphics[width=0.8\linewidth]{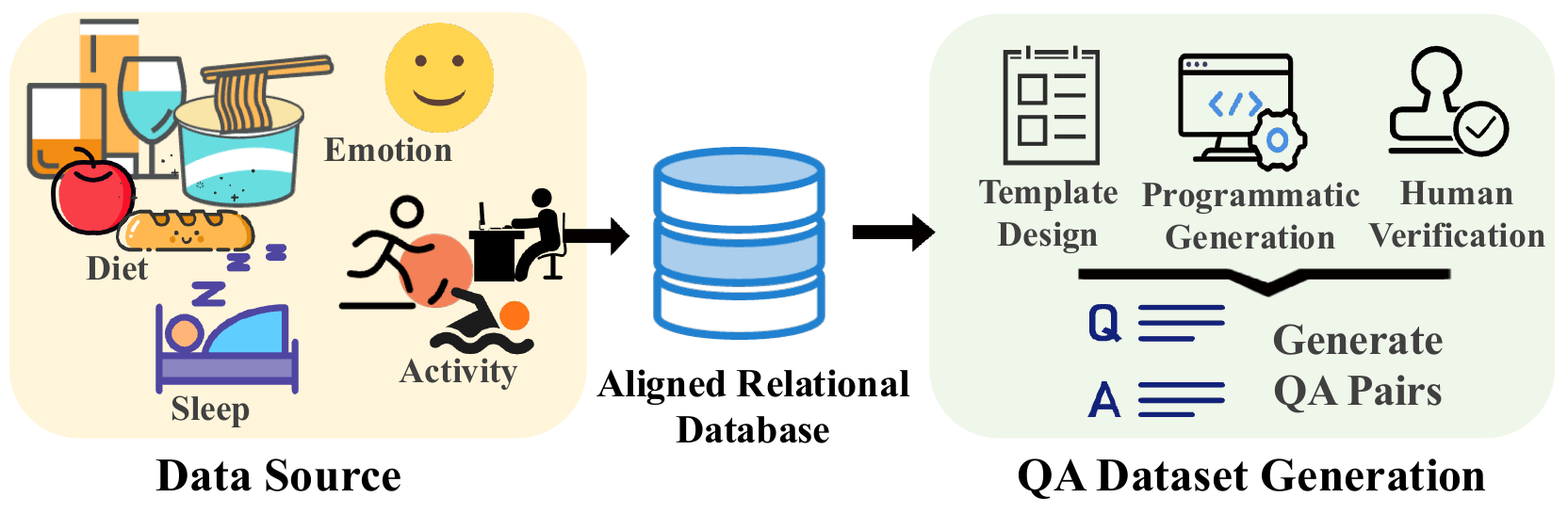}
%\vspace{-0.3cm}
\caption{Overview of the \systemname construction pipeline. Multi-dimensional lifestyle records, including diet, sleep, activity, and emotion, are first aligned by user IDs and timestamps into a relational database. Based on the aligned database, we design question templates, programmatically generate executable QA instances, and perform human verification to obtain the final QA pairs.}
\label{fig:DataGenerationPipline}
%\vspace{-0.4cm}
\end{figure*}

\section*{Appendix}

% \section{Reproducibility statement}

% To ensure full reproducibility of our results, we release all code for data generation, prompt construction, and evaluation, together with processed datasets, schema definitions and guidelines at \href{https://anonymous.4open.science/r/MultilifeQA-05D2}{https://anonymous.4open.science/r/MultilifeQA-05D2}. 
% Our experimental settings, including model configurations, decoding parameters, and hardware environments, are described in detail in Appendix~\ref{app:implementation-details}. 
% The exact prompt templates used in each evaluation setting are provided in Appendix~\ref{app:prompts}. 
% We also specify the evaluation metrics and correctness criteria in the main text. 
% With these resources, independent researchers can replicate our experiments and verify all reported numbers.

% \section{The Use of Large Language Models (LLMs)}

% We used LLMs to assist with grammar correction and polishing of the manuscript. All generated text was carefully checked and validated by the authors, who take full responsibility for the final content.

\section{Dataset Construction Details}\label{app:pipline}

Figure~\ref{fig:DataGenerationPipline} illustrates the overall construction pipeline of \systemnamenospace. 
Starting from heterogeneous lifestyle records, we first normalize and align multi-domain data into a relational database. 
We then design reusable question templates and instantiate them through executable programs to generate natural-language questions and deterministic answers. 
Finally, we apply programmatic checks and human verification to ensure the validity of the generated QA pairs. The dataset, construction pipeline, and evaluation code are publicly available at \url{https://gdfwj.github.io/LifeAgentBench/}.

\subsection{Dataset Construction Pipeline}\label{app:dataset_pipeline}

To systematically evaluate LLMs on long-horizon and cross-domain lifestyle health reasoning, we develop an automated and extensible QA construction pipeline. 
The pipeline converts structured lifestyle records into verifiable QA pairs through the following stages.

\paragraph{Data Normalization and Database Construction.}
We start from the structured records in AI4FoodDB and organize the data
into four health-related lifestyle domains: diet, sleep, physical
activity, and emotion.
The raw files are normalized into relational tables with consistent
attribute names, data types, user identifiers, and timestamp formats.
During this process, we also handle missing and anomalous records before
QA generation.
For indicators such as step count and aerobic activity, missing values
are set to zero, while missing values for continuous indicators such as
sleep duration are imputed using the mean of the user's valid records
within the corresponding week.
Clearly anomalous or invalid records are removed before the normalized
tables are used for benchmark construction.
Records from different domains are aligned by user IDs and timestamps,
which allows questions to combine evidence across multiple lifestyle
dimensions and temporal windows.
The normalized tables are then loaded into a MySQL database, enabling
both single-domain retrieval and cross-domain reasoning through
executable SQL queries.

\paragraph{Template Design.}
We design a set of reusable question templates to cover the major reasoning types in \systemnamenospace, including fact queries, aggregated statistics, numeric comparison, conditional queries, and trend analysis. 
Each template specifies the natural-language question pattern, required domains and tables, input parameters, answer format, and the reasoning operators needed to derive the answer. 
These operators include time-window selection, temporal alignment, aggregation, comparison, threshold-based filtering, and trend or consecutiveness detection. 
This template-based design allows the benchmark to cover diverse reasoning tasks while keeping each question executable and interpretable.

\paragraph{Programmatic QA Generation.}
Given a template, the pipeline instantiates concrete questions by sampling valid users, time windows, domains, thresholds, and other task-specific parameters from the database. 
For each instantiated question, the corresponding executable SQL query or program is used to retrieve the required evidence and derive the deterministic ground-truth answer. 
The generated instance contains the natural-language question, the answer, the answer type, the involved domains, the user scope, and the metadata needed for evaluation. 
This procedure maximizes the reliability of the ground-truth answers by deriving them directly from executable queries over the underlying lifestyle records, rather than relying on manual annotation or model-generated answers.

\paragraph{Quality Control and Human Verification.}
We apply programmatic checks to filter invalid or ambiguous QA instances. 
These checks verify that the executable query runs successfully, the returned evidence is non-empty when required, the derived answer matches the expected answer format, and the question metadata is consistent with the instantiated template. 
In addition, we conduct human verification to inspect generated questions, answer derivations, and representative examples from different reasoning categories. 
This step helps ensure that the generated QA pairs are semantically clear, executable, and aligned with the intended reasoning tasks.

\subsection{Extensibility of the Template Framework}
\label{app:template_extensibility}

A key design goal of \systemname is to provide not only a fixed benchmark
but also a reusable framework for constructing new benchmarks over
longitudinal lifestyle data.
The construction pipeline is modular, separating dataset-specific data
organization from question-template definition, executable answer
generation, and evaluation.
We release the complete construction code to facilitate such extensions
and encourage researchers to adapt the framework to new datasets,
lifestyle domains, and reasoning settings.

\paragraph{Adding New Data Sources.}
To extend the framework to a new dataset, researchers can first organize
the target records into structured tables with explicit user identifiers,
timestamps, and relevant attributes, and load them into the relational
database.
Dataset-specific fields and table schemas can then be mapped to the
variables required by the question templates by modifying the data
loading and schema-mapping configurations.
When the target dataset provides semantically compatible attributes,
the existing templates and executable generation pipeline can be reused
without redesigning the underlying reasoning tasks.

\paragraph{Defining and Extending Question Templates.}
The framework also allows researchers to customize the reasoning tasks
themselves.
Each template specifies a natural-language question pattern, parameter
space, required data fields, executable retrieval or computation logic,
answer derivation rule, and expected output format.
Researchers can instantiate new questions within the existing reasoning
categories, including Fact Query (FQ), Aggregated Statistics (AS),
Conditional Query (CQ), Numeric Comparison (NC), and Trend Analysis (TA).
New questions can be created by defining new attributes, conditions, or
parameter combinations.
They can also introduce new templates or reasoning categories for
settings not covered by the current benchmark, such as multi-hop
reasoning, personalized goal tracking, or longer-term behavioral change
analysis.
In this way, the framework can be extended at both the data level and
the reasoning-task level.

\paragraph{Generating and Evaluating New QA Pairs.}
Once the required schemas and templates are specified, the same
generation pipeline can instantiate questions, execute the corresponding
SQL queries or programs, and derive deterministic ground-truth answers.
The resulting QA pairs can be stored in the same format as
\systemnamenospace and evaluated using the existing Context Prompting
and Database-augmented Prompting protocols and correctness criteria.
This allows newly constructed benchmark variants to retain the same
executable and verifiable evaluation framework while incorporating new
data sources and reasoning challenges.

\paragraph{Cross-Dataset Example.}
To provide a concrete example of this extensibility, we apply the
framework to LifeSnaps~\cite{yfantidou2022lifesnaps}, a longitudinal
lifestyle dataset whose data structure and indicator organization differ
from those of AI4FoodDB.
Using the sleep records in LifeSnaps, we organize the relevant fields in
the relational database and map the dataset-specific attributes to the
corresponding semantic variables and template parameters.
We then reuse the existing template instantiation, executable query
generation, and deterministic answer-validation pipeline.
This adaptation generates 2,500 verifiable QA pairs covering all five
reasoning types: FQ, AS, CQ, NC, and TA.
The example provides empirical evidence that the construction framework
is not tied to the specific schema or study setting of AI4FoodDB.
More broadly, researchers can follow the same procedure to construct
benchmark variants over other longitudinal datasets, while defining new
mappings, domains, or executable reasoning templates when required by
the target data and research questions.

\begin{table}[t]
\centering
\small
\resizebox{\linewidth}{!}{%
\begin{tabular}{llrr}
\toprule
& Category & Count & Percentage \\
\midrule
User scope 
& Single-user & 13,452 & 59.6\% \\
& Multi-user  & 9,121  & 40.4\% \\
& Total       & 22,573 & 100.0\% \\
\midrule
Reasoning type 
& Fact Query (FQ) & 3,402 & 15.1\% \\
& Aggregated Statistics (AS) & 4,732 & 21.0\% \\
& Conditional Query (CQ) & 5,134 & 22.7\% \\
& Numeric Comparison (NC) & 4,550 & 20.2\% \\
& Trend Analysis (TA) & 4,755 & 21.1\% \\
\midrule
Answer format 
& Yes/No & 3,500 & 15.5\% \\
& Single Number & 10,642 & 47.1\% \\
& Short Text & 4,061 & 18.0\% \\
& Pairwise Answer & 3,186 & 14.1\% \\
& Multi-item Answer & 1,184 & 5.2\% \\
\midrule
Domain coverage 
& Activity & 10,003 & 44.3\% \\
& Sleep & 12,422 & 55.0\% \\
& Emotion & 6,950 & 30.8\% \\
& Diet & 6,213 & 27.5\% \\
\bottomrule
\end{tabular}%
}
\caption{Overall statistics of \systemnamenospace. Percentages are computed over all 22,573 QA pairs. Domain coverage is non-exclusive: a question involving multiple domains is counted once for each involved domain.}
\label{tab:dataset_summary}
\end{table}

\section{Dataset Statistics}\label{app:dataset_statistics}

Table~\ref{tab:dataset_summary} provides a compact summary of \systemnamenospace. 
The benchmark contains 22,573 QA pairs, including 13,452 single-user questions and 9,121 multi-user questions, covering both individualized lifestyle reasoning and multi-user comparison.
For reasoning types, the dataset is relatively balanced. 
Fact queries account for 15.1\% of the questions, while the other four reasoning types each account for about 20\%: aggregated statistics (21.0\%), conditional queries (22.7\%), numeric comparison (20.2\%), and trend analysis (21.1\%). 
This distribution allows the benchmark to evaluate a broad spectrum of reasoning capabilities. 
In addition to basic information retrieval, models are required to aggregate lifestyle signals over long time windows, apply threshold- or condition-based filtering, compare measurements across periods or users, and identify temporal patterns such as trends or consecutive changes.

The benchmark also covers diverse answer formats. Single-number answers account for the largest proportion (47.1\%). 
This reflects the fact that many lifestyle records are naturally represented as measurable quantities, such as duration, frequency, intensity, averages, and differences over time. 
However, \systemnamenospace is not limited to scalar prediction. 
It also includes yes/no answers (15.5\%), short-text answers (18.0\%), pairwise answers (14.1\%), and multi-item answers (5.2\%), which introduce additional requirements beyond computing a single value. 
These formats require models to make binary judgments, extract categorical information, associate values with dates or entities, and synthesize multiple related outputs in a structured form.

For domain coverage, the questions span four lifestyle dimensions: diet, sleep, physical activity, and emotion. 
Sleep and activity appear most frequently, while emotion and diet provide additional cross-domain evidence for more comprehensive lifestyle health reasoning. 
Since a question may involve multiple domains, domain coverage is non-exclusive: a cross-domain question is counted once for each involved domain. 
Overall, these statistics show that \systemname provides a diverse testbed for evaluating LLMs on long-horizon, cross-dimensional, and multi-user lifestyle health reasoning.

\section{Additional Experimental Results}\label{app:additional_results}

\subsection{Question Type and Answer Format}
Table~\ref{tab:all_res_by_qs_ans} provides a fine-grained breakdown of model performance by question type and answer format. 
The results reveal that the difficulty of \systemname is not uniform across categories: models are relatively more reliable on simple retrieval and constrained-output questions, but degrade substantially when the task requires long-window aggregation, numerical comparison, or structured multi-part answer generation.
Among question types, the contrast between Fact Query (FQ) and Aggregated Statistics (AS) highlights how performance changes when the task moves from record-level grounding to long-window evidence consolidation. 
FQ evaluates whether models can identify and interpret relevant lifestyle records, while AS keeps this grounding requirement but further requires aggregating multiple records over a temporal window. 
The results indicate that the aggregation requirement substantially increases the reasoning difficulty and leads to a clear drop in accuracy.
For example, under Context Prompting, Qwen2.5-7B obtains 19.38\% accuracy on FQ but only 5.21\% on AS, while GPT-4o drops from 69.78\% on FQ to 15.63\% on AS. 
This suggests that directly prompting LLMs with tabular context is not reliable enough for long-window aggregation.
Numeric Comparison (NC) introduces a related but distinct challenge: models must first derive or retrieve intermediate quantities and then compare them across time periods, users, or lifestyle dimensions. 
Under Database-augmented Prompting, GPT-4o achieves 87.52\% Acc|EX but only 31.31\% raw accuracy on NC. 
This gap indicates that once effective evidence is retrieved, the model can often derive the correct comparison, but generating the correct retrieval query remains a major bottleneck.

The answer-format breakdown further shows that models become less reliable when the answer requires more multi-step reasoning and synthesis.
Models perform better on constrained formats such as Yes/No and single-number answers, where the final response is short and usually centered on one decision or one scalar value. 
For instance, GPT-4o achieves 75.06\% accuracy on Yes/No answers and 68.31\% on single-number answers under Context Prompting, but its accuracy drops to 33.87\% on pairwise answers and 28.11\% on multi-item answers. 
This decline suggests that current models face a clear bottleneck when answering more reasoning-intensive multi-part questions. 
Unlike single-value outputs, pairwise and multi-item answers require the model to derive multiple pieces of evidence, maintain the logical correspondence among them, and synthesize them into a coherent structured response.
The gap between raw DP accuracy and Acc|EX further helps separate evidence retrieval from answer synthesis. 
For example, GPT-4o reaches 98.89\% Acc|EX on AS and 99.26\% on Conditional Query (CQ), while its corresponding raw DP accuracies are 26.45\% and 58.47\%. 
This shows that when the retrieved evidence is effective, strong models can often produce the correct final answer; however, generating the correct retrieval query remains difficult. 
Overall, these fine-grained results support the main finding that current LLMs struggle most with long-horizon aggregation, numerical comparison, and structured answer synthesis.

\subsection{User Scope and Domain Complexity}
Table~\ref{tab:merged_table_type} reports model performance under different user scopes and domain-complexity settings. 
The Single setting corresponds to questions grounded in one lifestyle dimension, while M-Sleep and M-Activity involve multiple types of indicators within the sleep and activity domains, respectively.
M-C2 and M-C4 further expand the evidence scope to cross-domain reasoning over two and four lifestyle dimensions. 
The multi-user setting additionally requires cohort-level aggregation or comparison across users.
The results show that expanding the evidence scope generally makes the task more challenging. 
For example, under Context Prompting, Qwen2.5-7B obtains 45.00\% accuracy in the Single setting, but drops to 14.69\% in the single-user M-C4 setting. 
A similar trend appears under Database-augmented Prompting, where Qwen2.5-7B decreases from 38.37\% on Single to only 2.81\% on single-user M-C4. 
Even stronger proprietary models show clear degradation when the evidence scope and domain coverage become more complex: GPT-4o obtains 55.67\% raw accuracy on Single under Database-augmented Prompting, but only 23.54\% on single-user M-C4. 
These results indicate that cross-domain reasoning over heterogeneous lifestyle records remains difficult, especially when the model must integrate evidence from multiple lifestyle domains and reason over the logical relationships among different health indicators.

The multi-user setting introduces another layer of difficulty because models must reason over cohort-level statistics rather than only an individual user's records. 
For instance, under Database-augmented Prompting, GPT-4o reaches 36.27\% accuracy on multi-user Single questions and 31.31\% on multi-user M-C4 questions, both lower than its single-user Single performance. 
This suggests that user-scope expansion increases the burden of evidence retrieval and aggregation, particularly when multi-user comparison is combined with cross-domain reasoning.
This results also shows that the model can usually use the retrieved evidence effectively once the relevant information is available, but the expanded domain scope makes it difficult to identify and retrieve the necessary evidence in the first place. 
Overall, these results show that both domain complexity and user-scope expansion substantially increase the difficulty of lifestyle health reasoning, highlighting the need for methods that can better integrate evidence across lifestyle domains, capture logical relationships among health indicators, and support cohort-level aggregation.

\subsection{Impact of Model Scale and Quantization}
Table~\ref{tab:qwen_results} compares Qwen2.5 variants across different model scales and quantization settings. 
The results show that model scale and numerical precision both affect performance, but the gains are not uniform across settings or reasoning categories.
Under Context Prompting, larger or higher-precision models generally improve overall accuracy. 
Qwen2.5-7B achieves 40.45\%, while Qwen2.5-14B (8-bit) and Qwen2.5-32B (4-bit) improve to 51.51\% and 53.86\%, respectively. 
However, scaling alone does not fully address difficult reasoning types: even Qwen2.5-32B (4-bit) reaches only 9.80\% accuracy on Aggregated Statistics (AS), indicating that long-window aggregation remains challenging.

Quantization introduces a clear trade-off. 
Although Qwen2.5-14B has more parameters than Qwen2.5-7B, the 4-bit version obtains only 38.42\% overall accuracy under Context Prompting, lower than the 7B model's 40.45\%, whereas the 8-bit version reaches 51.51\%. 
A similar gap appears under Database-augmented Prompting, where Qwen2.5-14B (4-bit) obtains 16.39\% raw accuracy and 54.82\% Acc|EX, while Qwen2.5-14B (8-bit) reaches 19.10\% and 77.82\%. 
These results suggest that aggressive quantization can weaken both evidence-based reasoning and final answer generation.
The Database-augmented Prompting results further show that larger models do not monotonically improve all metrics. 
Qwen2.5-32B (4-bit) achieves the highest raw DP accuracy among the Qwen variants (26.95\%), but its Acc|EX (75.24\%) is lower than that of Qwen2.5-7B (84.78\%). 
Overall, scaling and precision help improve performance, but they do not eliminate the core bottlenecks in long-horizon aggregation, structured answer synthesis, and cross-domain evidence grounding.

\subsection{Multi-Judge Evaluation of the Case Study}
\label{sec:multijudge}

To further examine the robustness of the case-study evaluation, we report the complete results from all five LLM judges in Table~\ref{tab:multijudge}.
The judges include Gemini-3.5-Flash~\cite{gemini35flash},
DeepSeek-V4-Pro~\cite{xu2026deepseek}, Claude-Sonnet-5~\cite{claudesonnet5}, Qwen3.5-9B~\cite{team2026qwen3}, and GPT-4o~\cite{openai2024gpt4o}.
This multi-judge protocol reduces dependence on any single LLM evaluator,
but does not substitute for human or health-expert validation and should
not be interpreted as clinical validation.
For the two open-ended tasks, all judges are provided with the same user task, supporting evidence, candidate responses, evaluation rubric, and 1--5 scoring scale, following the evaluation protocol described in Section~\ref{sec:case_study}.
Although the absolute scores vary across judges, the relative performance is highly consistent.
For every judge, \agentmname achieves the highest average score on both comprehensive health reasoning and targeted lifestyle recommendations.
The same trend also holds across the individual lifestyle dimensions reported in Table~\ref{tab:multijudge}.
These results indicate that the main case-study conclusion is robust to the choice of LLM judge rather than being driven by a single evaluator.

\section{The Use of Large Language Models (LLMs)}

We used LLM-based writing assistants for grammar correction, wording refinement, and improving the readability of the manuscript. 
All LLM-assisted edits were carefully reviewed, revised when necessary, and validated by the authors, who take full responsibility for the final content of the paper.

\section{Implementation Details}
\label{app:experimental-details}

\paragraph{Models and computing infrastructure.}
We evaluate both open-source and proprietary LLMs. 
For open-source models, including DeepSeek-Coder-1.3B, Llama-3.2-3B, Phi-3.5-mini-3.8B, Mistral-v0.3-7B, Qwen2.5-7B, Llama-3.1-8B, and Gemma-2-9B, Qwen3.5-9B are evaluated on a single RTX~4090 GPU. 
Larger models, including Qwen2.5-14B with 4-bit and 8-bit quantization, Qwen2.5-32B with 4-bit quantization, and Llama-3.1-70B with 4-bit quantization, are evaluated on 4$\times$A6000 GPUs. 
Closed-source models, including GPT-4o, Claude-3-Haiku, and Gemini 2.5 Flash-Lite, DeepSeek-V4-Pro are accessed through their official APIs. 
The total open-source inference cost was approximately 300--500 GPU hours, estimated from wall-clock runtime and the number of GPUs used.

\paragraph{Inference settings.}
All experiments are inference-only; we do not train or fine-tune any LLMs. 
We use a maximum context length of 4096 tokens, or the maximum supported length if it is smaller. 
For \emph{Context Prompting}, we set \texttt{max\_new\_tokens}=32. 
For \emph{Database-augmented Prompting}, we set \texttt{max\_new\_tokens}=480 for the SQL generation stage and \texttt{max\_new\_tokens}=48 for the final answer generation stage. 
For open-source models, we use deterministic decoding with temperature disabled, equivalent to greedy decoding. 
For closed-source models, we use deterministic or the lowest-temperature decoding setting supported by the corresponding API. 
No hyperparameter search is performed; the decoding and token-budget settings are fixed across models within each evaluation setting.

\section{Artifact License and Terms of Use.}
The released benchmark, construction pipeline, and evaluation code are provided for research use under non-commercial usage restrictions. The benchmark is derived from an existing research dataset, and its redistribution follows the terms of the original data source. License and usage details are included in the public repository.

%%%%%%%%%%%%%%%%%%%%%%%%%%%%%%%%%%%%%%%%%%%%%
\begin{table*}[t]
\centering
\newcolumntype{K}{wc{1.35cm}}
\resizebox{\linewidth}{!}{%
\begin{tabular}{l l | K K K | K K K}
\toprule
\textbf{Model} & \textbf{Dimension}
& \multicolumn{3}{c|}{\textbf{Task 2: Comprehensive Health Reasoning}}
& \multicolumn{3}{c}{\textbf{Task 3: Targeted Lifestyle Recommendations}} \\
\cmidrule(lr){3-5}
\cmidrule(lr){6-8}
& & \textbf{CP} & \textbf{DP} & \textbf{LifeAgent}
& \textbf{CP} & \textbf{DP} & \textbf{LifeAgent} \\
\midrule

% ================= Gemini =================
\multirow{6}{*}{Gemini-3.5-Flash}
& Activity       & 2.19 & 2.64 & 3.95 & 3.65 & 3.85 & 4.13 \\
& Sleep          & 2.42 & 2.54 & 3.41 & 3.96 & 4.28 & 4.44 \\
& Emotion        & 2.38 & 2.92 & 3.71 & 3.81 & 3.80 & 4.26 \\
& Diet           & 1.58 & 1.87 & 3.14 & 3.77 & 3.78 & 4.29 \\
& All dimensions & 3.31 & 3.86 & 4.21 & 3.46 & 3.55 & 3.85 \\
\rowcolor{gray!15}
& \textbf{Average}
& 2.38 & 2.77 & \textbf{3.68}
& 3.73 & 3.85 & \textbf{4.19} \\
\midrule

% ================= DeepSeek =================
\multirow{6}{*}{DeepSeek-V4-Pro}
& Activity       & 1.63 & 1.96 & 2.66 & 2.31 & 2.55 & 2.75 \\
& Sleep          & 1.90 & 2.06 & 2.91 & 2.63 & 2.83 & 3.20 \\
& Emotion        & 2.01 & 2.28 & 3.52 & 2.45 & 2.98 & 3.09 \\
& Diet           & 1.41 & 1.50 & 3.11 & 1.83 & 2.00 & 2.34 \\
& All dimensions & 2.21 & 2.37 & 2.63 & 1.89 & 2.02 & 2.25 \\
\rowcolor{gray!15}
& \textbf{Average}
& 1.83 & 2.03 & \textbf{2.97}
& 2.22 & 2.48 & \textbf{2.73} \\
\midrule

% ================= Claude =================
\multirow{6}{*}{Claude-Sonnet-5}
& Activity       & 1.59 & 2.22 & 3.37 & 1.68 & 1.93 & 2.26 \\
& Sleep          & 1.84 & 2.10 & 3.12 & 2.40 & 2.72 & 2.84 \\
& Emotion        & 1.69 & 2.31 & 3.52 & 2.12 & 2.27 & 2.67 \\
& Diet           & 1.25 & 1.33 & 3.34 & 2.05 & 2.28 & 2.47 \\
& All dimensions & 1.94 & 2.37 & 2.49 & 1.75 & 1.82 & 2.25 \\
\rowcolor{gray!15}
& \textbf{Average}
& 1.66 & 2.07 & \textbf{3.17}
& 2.00 & 2.20 & \textbf{2.50} \\
\midrule

% ================= Qwen =================
\multirow{6}{*}{Qwen3.5-9B}
& Activity       & 1.50 & 2.05 & 3.08 & 1.73 & 1.97 & 2.30 \\
& Sleep          & 2.46 & 2.46 & 3.36 & 3.16 & 3.65 & 3.97 \\
& Emotion        & 2.14 & 2.52 & 3.64 & 1.86 & 2.09 & 2.50 \\
& Diet           & 1.27 & 1.27 & 2.94 & 2.15 & 2.45 & 2.96 \\
& All dimensions & 1.70 & 1.93 & 2.07 & 1.05 & 1.22 & 1.56 \\
\rowcolor{gray!15}
& \textbf{Average}
& 1.81 & 2.05 & \textbf{3.02}
& 1.99 & 2.28 & \textbf{2.66} \\
\midrule

% ================= GPT-4o =================
\multirow{6}{*}{GPT-4o}
& Activity       & 1.23 & 1.54 & 2.76 & 3.73 & 3.91 & 3.99 \\
& Sleep          & 1.45 & 1.38 & 2.47 & 4.16 & 4.30 & 4.39 \\
& Emotion        & 1.21 & 1.66 & 2.63 & 3.20 & 3.42 & 3.69 \\
& Diet           & 1.11 & 1.12 & 2.61 & 3.22 & 3.29 & 3.54 \\
& All dimensions & 1.53 & 1.79 & 2.08 & 4.40 & 4.16 & 4.43 \\
\rowcolor{gray!15}
& \textbf{Average}
& 1.31 & 1.50 & \textbf{2.51}
& 3.74 & 3.82 & \textbf{4.01} \\
\bottomrule
\end{tabular}%
}
\caption{Detailed multi-judge evaluation results for the two open-ended case-study tasks.
All judges use the same evaluation rubric and 1--5 scoring scale.
Gray rows report the average scores for each judge, with the best average score for each task shown in bold.}
\label{tab:multijudge}
\end{table*}
%%%%%%%%%%%%%%%%%%%%%%%%%%%%%%%%%%%%%%%%%%%%%

\begin{table*}[h]
\centering
\resizebox{\textwidth}{!}{
\setlength{\tabcolsep}{4pt}
\begin{tabular}{l|l|l|ccccc|ccccc}
\toprule
\multicolumn{3}{c|}{} & \multicolumn{5}{c|}{\textbf{Question Type}} & \multicolumn{5}{c}{\textbf{Answer Type}} \\
\cmidrule(lr){4-8}\cmidrule(lr){9-13}
\textbf{Model} & \textbf{Setup} & \textbf{Metric} & FQ & AS & CQ & NC & TA & Yes/No & Single Number & Short Text & Pairwise Answer & Multi-item Answer \\
\midrule
\multicolumn{13}{c}{\textbf{Open-source LLMs}} \\
\midrule
\rowcolor{gray!15}\multirow{3}{*}{deepseek-coder-1.3B} & Context Prompting & Acc   & 0.47 & 0.04 & 0.0 & 0.37 & 4.78 & 8.57 & 0.32 & 0.76 & 0.0 & 0.0 \\
  & \multirow{2}{*}{Database-augmented Prompting} & Acc   & 0.06 & 0.08 & 2.90 & 1.60 & 1.18 & 3.00 & 1.68 & 0.0 & 0.0 & 0.0 \\
  &                              & Acc|EX & 0.21 & 0.20 & 2.15 & 10.55 & 11.95 & 14.54 & 1.76 & 0.0 & 0.0 & 0.0 \\
\midrule
   \rowcolor{gray!15}\multirow{3}{*}{Llama-3.2-3B} & Context Prompting & Acc   & 10.27 & 1.17 & 38.95 & 10.75 & 34.82 & 33.06 & 29.92 & 21.33 & 0.98 & 0.09 \\
  & \multirow{2}{*}{Database-augmented Prompting} & Acc   & 17.72 & 9.82 & 19.81 & 11.63 & 9.00 & 17.78 & 16.04 & 17.43 & 0.13 & 0.0 \\
  &                              & Acc|EX & 68.69 & 84.64 & 79.19 & 52.87 & 52.32 & 52.05 & 81.19 & 69.60 & 6.67 & 0.0 \\
\midrule
   \rowcolor{gray!15}\multirow{3}{*}{Phi-3.5-mini-3.8B} & Context Prompting & Acc   & 13.96 & 2.53 & 28.42 & 19.96 & 35.10 & 41.62 & 27.70 & 23.70 & 1.84 & 0.09 \\
  & \multirow{2}{*}{Database-augmented Prompting} & Acc   & 25.10 & 10.65 & 24.68 & 13.01 & 9.06 & 9.37 & 21.43 & 25.46 & 0.16 & 0.0 \\
  &                              & Acc|EX & 69.09 & 91.94 & 93.95 & 65.98 & 61.87 & 51.17 & 90.33 & 86.94 & 1.73 & 0.0 \\
\midrule
   \rowcolor{gray!15}\multirow{3}{*}{Mistral-v0.3-7B} & Context Prompting & Acc   & 24.47 & 2.83 & 49.43 & 13.89 & 59.69 & 48.66 & 44.61 & 28.13 & 6.79 & 0.89 \\
  & \multirow{2}{*}{Database-augmented Prompting} & Acc   & 20.99 & 6.78 & 11.24 & 9.01 & 0.34 & 8.63 & 10.75 & 14.55 & 0.03 & 0.0 \\
  &                              & Acc|EX & 82.45 & 77.48 & 81.69 & 59.76 & 55.00 & 55.21 & 83.92 & 88.50 & 1.96 & 0.0 \\
\midrule
   \rowcolor{gray!15}\multirow{3}{*}{Qwen2.5-7B} & Context Prompting & Acc   & 19.38 & 5.21 & 70.34 & 24.42 & 73.56 & 63.38 & 60.69 & 42.66 & 1.78 & 0.36 \\
  & \multirow{2}{*}{Database-augmented Prompting} & Acc   & 29.78 & 17.37 & 36.21 & 12.90 & 11.80 & 11.71 & 27.07 & 32.55 & 6.06 & 3.04 \\
  &                              & Acc|EX & 83.57 & 94.19 & 95.15 & 65.86 & 70.03 & 60.21 & 97.53 & 80.05 & 63.46 & 29.51 \\
\midrule
   \rowcolor{gray!15}\multirow{3}{*}{Llama-3.1-8B} & Context Prompting & Acc   & 20.93 & 2.83 & 24.67 & 17.92 & 36.03 & 56.78 & 21.31 & 25.31 & 5.24 & 4.45 \\
  & \multirow{2}{*}{Database-augmented Prompting} & Acc   & 24.96 & 11.39 & 30.46 & 28.13 & 13.21 & 24.68 & 25.93 & 28.27 & 1.51 & 3.46 \\
  &                              & Acc|EX & 67.44 & 82.74 & 97.77 & 71.32 & 66.27 & 58.14 & 97.35 & 82.89 & 14.33 & 18.46 \\
\midrule
   \rowcolor{gray!15}\multirow{3}{*}{gemma-2-IT-9B} & Context Prompting & Acc   & 23.90 & 4.29 & 33.18 & 14.15 & 44.87 & 22.44 & 35.70 & 37.92 & 6.00 & 0.0 \\
  & \multirow{2}{*}{Database-augmented Prompting} & Acc   & 19.08 & 9.62 & 17.04 & 17.43 & 10.71 & 15.97 & 15.64 & 24.43 & 1.69 & 1.01 \\
  &                              & Acc|EX & 59.43 & 51.15 & 55.78 & 47.13 & 41.25 & 35.00 & 58.05 & 66.60 & 18.93 & 7.69 \\
\midrule
\rowcolor{gray!15}\multirow{3}{*}{Qwen2.5-14B (4-bit)} &  Context Prompting & Acc   & 17.49 & 6.24 & 62.97 & 30.24 & 71.45 & 56.39 & 59.56 & 49.84 & 0.35 & 0.18 \\
  & \multirow{2}{*}{Database-augmented Prompting} & Acc   & 22.87 & 13.10 & 22.38 & 15.78 & 9.13 & 11.60 & 16.89 & 30.16 & 6.94 & 4.22  \\
  &                              & Acc|EX & 61.01 & 60.59 & 50.10 & 49.20 & 63.16 & 49.21 & 50.40 & 68.91 & 58.75 & 27.22 \\
\midrule
\rowcolor{gray!15}\multirow{3}{*}{Qwen2.5-14B (8-bit)} & Context Prompting & Acc   & 60.54 & 8.29 & 75.89 & 35.65 & 75.52 & 62.72 & 66.97 & 52.14 & 26.19 & 19.48 \\
  & \multirow{2}{*}{Database-augmented Prompting} & Acc   & 29.59 & 17.05 & 29.59 & 18.31 & 5.78 & 9.26 & 22.35 & 33.02 & 7.31 & 2.87  \\
  &                              & Acc|EX & 84.23 & 78.30 & 84.23 & 74.68 & 72.30 & 62.19 & 82.44 & 83.56 & 63.66 & 21.85 \\
\midrule
\rowcolor{gray!15}\multirow{3}{*}{Qwen2.5-32B (4-bit)} & Context Prompting & Acc   & 65.48 & 9.80 & 69.98 & 41.94 & 81.90 & 71.99 & 67.46 & 56.12 & 28.38 & 21.17 \\
  & \multirow{2}{*}{Database-augmented Prompting} & Acc   & 29.75 & 19.65 & 46.67 & 25.49 & 12.34 & 17.71 & 34.03 & 37.85 & 8.29 & 3.63  \\
  &                              & Acc|EX & 64.84 & 79.84 & 79.71 & 73.18 & 76.50 & 69.70 & 79.19 & 82.54 & 48.95 & 20.28 \\
\midrule
   \rowcolor{gray!15}\multirow{3}{*}{Llama-3.1-70B} & Context Prompting & Acc   & 47.52 & 8.00 & 40.10 & 34.59 & 73.02 & 65.15 & 48.16 & 49.92 & 19.81 & 10.41 \\
  & \multirow{2}{*}{Database-augmented Prompting} & Acc   & 15.26 & 7.78 & 22.22 & 16.55 & 7.57 & 11.80 & 15.28 & 26.08 & 1.35 & 0.0 \\
  &                              & Acc|EX & 52.43 & 56.29 & 67.73 & 54.93 & 51.04 & 46.30 & 63.79 & 74.90 & 12.91 & 0.0 \\
\midrule
\multicolumn{13}{c}{\textbf{Closed-source LLMs}} \\
\midrule
   \rowcolor{gray!15}\multirow{3}{*}{Gemini 2.5 Flash-Lite} & Context Prompting & Acc   & 59.74 & 9.80 & 54.78 & 33.42 & 67.74 & 48.85 & 56.56 & 52.68 & 25.71 & 17.62 \\
  & \multirow{2}{*}{Database-augmented Prompting} & Acc   & 39.04 & 34.86 & 54.32 & 44.46 & 36.23 & 44.89 & 45.49 & 50.26 & 9.26 & 5.41 \\
  &                              & Acc|EX & 82.92 & 73.75 & 98.30 & 76.89 & 78.88 & 70.23 & 97.92 & 84.75 & 40.31 & 22.86\\
\midrule
   \rowcolor{gray!15}\multirow{3}{*}{Claude-3-haiku} & Context Prompting & Acc   & 45.73 & 7.01 & 52.20 & 23.95 & 47.51 & 41.05 & 46.29 & 39.98 & 17.27 & 10.77 \\
  & \multirow{2}{*}{Database-augmented Prompting} & Acc   & 28.22 & 15.49 & 45.62 & 29.01 & 26.46 & 34.31 & 32.78 & 44.74 & 1.76 & 4.31 \\
  &                              & Acc|EX & 64.95 & 79.12 & 98.35 & 64.25 & 63.88 & 56.73 & 94.40 & 90.28 & 9.56 & 21.25 \\
\midrule
   \rowcolor{gray!15}\multirow{3}{*}{GPT-4o} & Context Prompting & Acc   & 69.78 & 15.63 & 71.45 & 49.58 & 79.08 & 75.06 & 68.31 & 61.62 & 33.87 & 28.11 \\
  & \multirow{2}{*}{Database-augmented Prompting} & Acc   & 42.50 & 26.45 & 58.47 & 31.31 & 14.89 & 19.29 & 41.50 & 50.41 & 15.63 & 16.98 \\
  &                              & Acc|EX & 93.74 & 98.89 & 99.26 & 87.52 & 96.85 & 77.59 & 99.23 & 99.17 & 86.04 & 87.78 \\
\bottomrule
\end{tabular}%
}
\caption{Fine-grained LLMs performance by question type and answer format. 
Question types include Fact Query (FQ), Aggregated Statistics (AS), Conditional Query (CQ), Numeric Comparison (NC), and Trend Analysis (TA). 
Answer formats include Yes/No, Single Number, Short Text, Pairwise Answer, and Multi-item Answer. 
}
\label{tab:all_res_by_qs_ans}
\end{table*}

\begin{table*}[t]
\centering
\resizebox{\textwidth}{!}{
\setlength{\tabcolsep}{4pt}
\begin{tabular}{l|l|l|ccccc|ccccc|cc}
\toprule
\multicolumn{3}{c|}{} & \multicolumn{5}{c|}{\textbf{Question Type}} & \multicolumn{5}{c|}{\textbf{Answer Type}} & \multirow{2}{*}{\textbf{Overall}}  \\
\cmidrule(lr){4-8}\cmidrule(lr){9-13}
\textbf{Model} & \textbf{Setup} & \textbf{Metric} & FQ & AS & CQ & NC & TA & Yes/No & Single Number & Short Text & Pairwise Answer & Multi-item Answer & \\
\midrule
 \rowcolor{gray!15}\multirow{3}{*}{Qwen2.5-7B} & Context Prompting & Acc   & 19.38 & 5.21 & 70.34 & 24.42 & 73.56 & 63.38 & 60.69 & 42.66 & 1.78 & 0.36 & 40.45 \\
  & \multirow{2}{*}{Database-augmented Prompting} & Acc   & 29.78 & 17.37 & 36.21 & 12.90 & 11.80 & 11.71 & 27.07 & 32.55 & 6.06 & 3.04 & 21.45 \\
  &                              & Acc|EX & 83.57 & 94.19 & 95.15 & 65.86 & 70.03 & 60.21 & 97.53 & 80.05 & 63.46 & 29.51 & 84.78 \\
\midrule
\rowcolor{gray!15}\multirow{3}{*}{Qwen2.5-14B (4-bit)} &  Context Prompting & Acc   & 17.49 & 6.24 & 62.97 & 30.24 & 71.45 & 56.39 & 59.56 & 49.84 & 0.35 & 0.18 & 38.42 \\
  & \multirow{2}{*}{Database-augmented Prompting} & Acc   & 22.87 & 13.10 & 22.38 & 15.78 & 9.13 & 11.60 & 16.89 & 30.16 & 6.94 & 4.22 & 16.39  \\
  &                              & Acc|EX & 61.01 & 60.59 & 50.10 & 49.20 & 63.16 & 49.21 & 50.40 & 68.91 & 58.75 & 27.22 & 54.82 \\
\midrule
\rowcolor{gray!15}\multirow{3}{*}{Qwen2.5-14B (8-bit)} & Context Prompting & Acc   & 60.54 & 8.29 & 75.89 & 35.65 & 75.52 & 62.72 & 66.97 & 52.14 & 26.19 & 19.48 & 51.51\\
  & \multirow{2}{*}{Database-augmented Prompting} & Acc   & 29.59 & 17.05 & 29.59 & 18.31 & 5.78 & 9.26 & 22.35 & 33.02 & 7.31 & 2.87 & 19.10 \\
  &                              & Acc|EX & 84.23 & 78.30 & 84.23 & 74.68 & 72.30 & 62.19 & 82.44 & 83.56 & 63.66 & 21.85 & 77.82 \\
\midrule
\rowcolor{gray!15}\multirow{3}{*}{Qwen2.5-32B (4-bit)} & Context Prompting & Acc   & 65.48 & 9.80 & 69.98 & 41.94 & 81.90 & 71.99 & 67.46 & 56.12 & 28.38 & 21.17 & 53.86\\
  & \multirow{2}{*}{Database-augmented Prompting} & Acc   & 29.75 & 19.65 & 46.67 & 25.49 & 12.34 & 17.71 & 34.03 & 37.85 & 8.29 & 3.63 & 26.95 \\
  &                              & Acc|EX & 64.84 & 79.84 & 79.71 & 73.18 & 76.50 & 69.70 & 79.19 & 82.54 & 48.95 & 20.28 & 75.24 \\
\bottomrule
\end{tabular}%
}
\caption{Performance of Qwen2.5 variants under different model scales and quantization settings. 
We report results by question type, answer format, and overall accuracy. }
\label{tab:qwen_results}
\end{table*}

\begin{table*}[t]
\centering
\resizebox{\textwidth}{!}{
\begin{tabular}{l|c|l|ccccc|cc}
\toprule
\multirow{2}{*}{Model} & \multirow{2}{*}{Setup}&\multirow{2}{*}{Metric} & \multicolumn{5}{c|}{Single-user} & \multicolumn{2}{c}{Multi-user} \\
\cmidrule(lr){4-8}\cmidrule(lr){9-10}
 &  &  & Single & M-Sleep & M-Activity & M-C2 & M-C4 & Single & M-C4 \\
\midrule
\multicolumn{10}{c}{\textbf{Open-source LLMs}} \\
\midrule
\rowcolor{gray!15}\multirow{3}{*}{deepseek-coder-1.3B} & Context Prompting & Acc   & 0.77 & 0.72 & 3.57 & 0.37 & 2.92 & \texttt{/} & \texttt{/} \\
& \multirow{2}{*}{Database-augmented Prompting} & Acc   & 2.90 & 2.88 & 0.64 & 1.05 & 0.00 & 1.01 & 0.00 \\
&  & Acc|EX & 2.39 & 21.05 & 1.59 & 5.95 & 0.00 & 4.17 & 0.00 \\
\midrule
\rowcolor{gray!15}\multirow{3}{*}{Llama-3.2-3B} & Context Prompting & Acc   & 22.53 & 10.07 & 23.68 & 16.90 & 29.17 & \texttt{/} & \texttt{/} \\
& \multirow{2}{*}{Database-augmented Prompting} & Acc   & 17.42 & 0.00 & 4.21 & 6.17 & 0.63 & 21.58 & 12.87 \\
&  & Acc|EX & 66.78 & 0.00 & 58.62 & 58.95 & 75.00 & 71.54 & 68.11 \\
\midrule
\rowcolor{gray!15}\multirow{3}{*}{Phi-3.5-mini-3.8B} & Context Prompting & Acc   & 28.10 & 14.75 & 26.67 & 17.27 & 2.08 & \texttt{/} & \texttt{/} \\
& \multirow{2}{*}{Database-augmented Prompting} & Acc   & 32.06 & 0.36 & 3.33 & 4.99 & 2.71 & 23.72 & 6.15 \\
&  & Acc|EX & 86.45 & 1.35 & 53.40 & 50.09 & 92.86 & 83.37 & 56.15 \\
\midrule
\rowcolor{gray!15}\multirow{3}{*}{Mistral-v0.3-7B} & Context Prompting & Acc   & 34.60 & 24.82 & 42.46 & 29.30 & 8.33 & \texttt{/} & \texttt{/} \\
& \multirow{2}{*}{Database-augmented Prompting} & Acc   & 19.71 & 0.00 & 0.70 & 0.45 & 0.31 & 15.15 & 0.25 \\
&  & Acc|EX & 77.57 & 0.00 & 100.00 & 18.84 & 100.00 & 76.34 & 75.00 \\
\midrule
\rowcolor{gray!15}\multirow{3}{*}{Qwen2.5-7B} & Context Prompting & Acc   & 45.00 & 18.70 & 48.54 & 40.23 & 14.69 & \texttt{/} & \texttt{/} \\
& \multirow{2}{*}{Database-augmented Prompting} & Acc   & 38.37 & 21.22 & 15.43 & 12.74 & 2.81 & 24.52 & 15.00 \\
&  & Acc|EX & 86.28 & 83.10 & 88.00 & 86.33 & 67.50 & 83.97 & 74.59 \\
\midrule
\rowcolor{gray!15}\multirow{3}{*}{Llama-3.1-8B} & Context Prompting & Acc   & 28.07 & 19.06 & 21.11 & 16.73 & 15.63 & \texttt{/} & \texttt{/} \\
& \multirow{2}{*}{Database-augmented Prompting} & Acc   & 35.15 & 4.68 & 8.95 & 12.91 & 4.48 & 27.12 & 20.98 \\
&  & Acc|EX & 91.01 & 13.27 & 60.96 & 68.29 & 48.31 & 80.43 & 69.32 \\
\midrule
\rowcolor{gray!15}\multirow{3}{*}{gemma-2-IT-9B} & Context Prompting & Acc   & 42.59 & 11.15 & 38.25 & 13.68 & 0.00 & \texttt{/} & \texttt{/} \\
& \multirow{2}{*}{Database-augmented Prompting} & Acc   & 22.01 & 4.32 & 6.08 & 7.27 & 3.02 & 21.26 & 9.03 \\
&  & Acc|EX & 53.27 & 28.57 & 32.10 & 34.46 & 26.61 & 61.74 & 47.74 \\
\midrule
\rowcolor{gray!15}\multirow{3}{*}{Qwen2.5-14B (4-bit)} & Context Prompting & Acc   & 52.54 & 14.75 & 45.44 & 34.89 & 11.35 & \texttt{/} & \texttt{/} \\
& \multirow{2}{*}{Database-augmented Prompting} & Acc   & 22.46 & 5.03 & 10.79 & 10.19 & 18.65 & 20.01 & 21.15 \\
&  & Acc|EX & 45.26 & 54.72 & 23.99 & 55.84 & 89.95 & 58.23 & 93.39 \\
\midrule
\rowcolor{gray!15}\multirow{3}{*}{Qwen2.5-14B (8-bit)} & Context Prompting & Acc   & 53.26 & 44.60 & 53.98 & 51.43 & 42.40 & \texttt{/} & \texttt{/} \\
& \multirow{2}{*}{Database-augmented Prompting} & Acc   & 36.01 & 4.68 & 3.16 & 9.22 & 10.63 & 24.00 & 16.23 \\
&  & Acc|EX & 85.04 & 31.71 & 26.70 & 82.70 & 26.70 & 75.09 & 94.33 \\
\midrule
\rowcolor{gray!15}\multirow{3}{*}{Qwen2.5-32B (4-bit)} & Context Prompting & Acc   & 59.09 & 48.92 & 54.80 & 51.14 & 50.10 & \texttt{/} & \texttt{/} \\
& \multirow{2}{*}{Database-augmented Prompting} & Acc   & 39.31 & 2.52 & 26.90 & 16.01 & 20.31 & 30.82 & 30.08 \\
&  & Acc|EX & 75.12 & 5.79 & 85.29 & 65.49 & 87.44 & 78.70 & 88.97 \\
\midrule
\rowcolor{gray!15}\multirow{3}{*}{Llama-3.1-70B (4-bit)} & Context Prompting & Acc   & 51.28 & 35.25 & 42.51 & 33.76 & 38.85 & \texttt{/} & \texttt{/} \\
& \multirow{2}{*}{Database-augmented Prompting} & Acc   & 16.95 & 0.72 & 4.39 & 7.29 & 12.71 & 20.39 & 14.43 \\
&  & Acc|EX & 54.18 & 11.76 & 24.15 & 46.18 & 62.56 & 69.19 & 73.13 \\
\midrule
\multicolumn{10}{c}{\textbf{Closed-source LLMs}} \\
\midrule
\rowcolor{gray!15}\multirow{3}{*}{Gemini 2.5 Flash-Lite} & Context Prompting & Acc   & 55.35 & 46.76 & 44.21 & 40.20 & 32.19 & \texttt{/} & \texttt{/} \\
& \multirow{2}{*}{Database-augmented Prompting} & Acc   & 54.13 & 18.71 & 25.38 & 32.09 & 22.81 & 42.40 & 40.81 \\
&  & Acc|EX & 93.47 & 41.60 & 80.00 & 77.86 & 60.07 & 84.10 & 78.25 \\
\midrule
\rowcolor{gray!15}\multirow{3}{*}{Claude-3-haiku} & Context Prompting & Acc   & 45.79 & 42.81 & 42.81 & 30.01 & 22.71 & \texttt{/} & \texttt{/} \\
& \multirow{2}{*}{Database-augmented Prompting} & Acc   & 39.54 & 11.15 & 23.74 & 23.14 & 12.19 & 33.73 & 19.51 \\
&  & Acc|EX & 80.68 & 28.85 & 69.28 & 62.91 & 55.39 & 89.02 & 70.21 \\
\midrule
\rowcolor{gray!15}\multirow{3}{*}{GPT-4o} & Context Prompting & Acc   & 62.64 & 47.12 & 54.67 & 54.51 & 57.39 & \texttt{/} & \texttt{/} \\
& \multirow{2}{*}{Database-augmented Prompting} & Acc   & 55.67 & 32.73 & 31.40 & 23.04 & 23.54 & 36.27 & 31.31 \\
&  & Acc|EX & 98.40 & 89.22 & 99.63 & 93.40 & 99.12 & 94.71 & 90.63 \\
\bottomrule
\end{tabular}
}
\caption{Performance by user scope and domain complexity. 
Single denotes questions grounded in one lifestyle dimension; M-Sleep and M-Activity denote questions involving multiple indicator types within the sleep and activity domains, respectively; M-C2 and M-C4 denote cross-domain questions involving two and four lifestyle dimensions.}
\label{tab:merged_table_type}
\end{table*}

\end{document}